\documentclass[nohyperref]{article}

\newcommand{\allFuncs}[1]{\mathbb{F}({#1})}
\newcommand{\tableScan}{\tau}
\newcommand{\commAssoc}{\oplus}
\newcommand{\grouping}{\texttt{grp}}
\newcommand{\joinPred}{\texttt{pred}}
\newcommand{\joinProj}{\texttt{proj}}
\newcommand{\selectPred}{\texttt{pred}}
\newcommand{\add}{\texttt{add}}
\newcommand{\selectKern}{\odot}
\newcommand{\selectProj}{\texttt{proj}}
\newcommand{\joinKern}{\otimes}

\newcommand{\boundellipse}[3]
{(#1) ellipse (#2 and #3)
}
\usepackage{microtype}
\usepackage{amsfonts}
\usepackage{booktabs} 
\usepackage[table,x11names]{xcolor}
\usepackage{hyperref}
\usepackage{balance}  
\usepackage{url}
\usepackage{xcolor}
\usepackage{listings}
\usepackage{makecell}
\usepackage{xcolor,cancel}
\usepackage{algorithm,algorithmic}
\usepackage{tikz}
\usepackage{float}
\usepackage{subfigure}
\usepackage{pifont}
\usepackage{blindtext}
\usepackage{amsmath, xparse,amsthm}
\usepackage{latexsym}
\usepackage{url}
\usepackage[capitalize,noabbrev]{cleveref}
\usepackage{graphicx}
\usepackage{caption}
\usetikzlibrary{calc,trees,positioning,arrows,fit,shapes,calc}

\theoremstyle{plain}

\theoremstyle{definition}

\theoremstyle{remark}

\usepackage[accepted]{icml2023}

\icmltitlerunning{Auto-Differentiation of Relational Computations for Very Large Scale Machine
Learning}

\lstnewenvironment{DSL}
  {\lstset{
        aboveskip=5pt,
        belowskip=5pt,
        escapechar=!,
        mathescape=true,
        language=SQL,
        basicstyle=\linespread{0.96}\ttfamily\small,
        morekeywords={
        Map, HashMap, aggregate, join, select, integer
        },
        deletekeywords={VALUE, PRIOR},
        showstringspaces=true}
        \vspace{0pt}%
        \noindent\minipage{0.47\textwidth}}
  {\endminipage\vspace{0pt}}
  
  \lstnewenvironment{SQL}
  {\lstset{
        aboveskip=5pt,
        belowskip=5pt,
        escapechar=!,
        mathescape=false,
        language=SQL,
        basicstyle=\linespread{0.94}\ttfamily\small,
        morekeywords={BULK, COLLECT, FOR, MATRIX, VECTOR, EACH, INTEGER, LONG, DOUBLE, WITH, PARTITION, TEST, WHETHER, PROBABILITY, VECTORIZE, ROWMATRIX, COLMATRIX, inner\_product, matrix\_vector\_multiply, MATRIX\_MULTIPLY, outer\_product, matrix\_inverse, trans\_matrix,
        label\_scalar, label\_vector, get\_scalar, get\_block, label\_matrix, diag, rowMins, rowIndexMax, map, transpose, multiply, asInstanceOf, DenseMatrix, 
        toArray, zipped, toIndexedRowMatrix, if, Tuple2, override, def, new, gemm, build, filter, matmul, crossentropyderiv, reluderiv, t, softmax, reducebyrow},
        deletekeywords={VALUE, PRIOR, Column, COLUMN},
        showstringspaces=true
}
\vspace{0pt}%
\noindent\minipage{0.47\textwidth}}
{\endminipage\vspace{0pt}}
\definecolor{codegreen}{rgb}{0,0.6,0}
\definecolor{codegray}{rgb}{0.5,0.5,0.5}
\definecolor{codepurple}{rgb}{0.58,0,0.82}
\definecolor{backcolour}{rgb}{1,1,1} 
\lstdefinestyle{mystyle}{
      backgroundcolor=\color{backcolour},   
      commentstyle=\color{codegreen},
      keywordstyle=\color{magenta},
      numberstyle=\tiny\color{codegray},
      stringstyle=\color{codepurple},
      basicstyle=\ttfamily\footnotesize,
      breakatwhitespace=false,         
      breaklines=true,                 
      captionpos=b,                    
      keepspaces=true,                 
      numbers=left,                    
      numbersep=5pt,                  
      showspaces=false,                
      showstringspaces=false,
      showtabs=false,                  
      tabsize=2
}
\begin{document}

\twocolumn[
\icmltitle{Auto-Differentiation of Relational Computations for Very Large Scale Machine Learning}




\begin{icmlauthorlist}
\icmlauthor{Yuxin Tang}{to}
\icmlauthor{Zhimin Ding}{to}
\icmlauthor{Dimitrije Jankov}{to}
\icmlauthor{Binhang Yuan}{ed}
\icmlauthor{Daniel Bourgeois}{to}
\icmlauthor{Chris Jermaine}{to}
\end{icmlauthorlist}

\icmlaffiliation{to}{Department of Computer Science, Rice University, Houston, US}
\icmlaffiliation{ed}{ETH Zurich, Switzerland}

\icmlcorrespondingauthor{Yuxin Tang}{yuxintang1995@gmail.com}
\icmlcorrespondingauthor{Chris Jermaine}{cmj4@rice.edu}

\icmlkeywords{Machine Learning, Auto Differentiation, ICML}

\vskip 0.3in
]



\printAffiliationsAndNotice{} 

\begin{abstract}
The relational data model was designed to facilitate large-scale data management and analytics. We consider the problem of how to differentiate computations expressed relationally.  We show experimentally that a relational engine running an auto-differentiated relational algorithm can easily scale to very large datasets, and is competitive with state-of-the-art, special-purpose systems for large-scale distributed machine learning.
\end{abstract}

\section{Introduction}
\label{submission}
The relational data model \cite{10.1145/362384.362685} is the basis for most modern SQL database systems. SQL can be used to extract and transform data into formats that can be used to train machine learning models. Furthermore, many data management systems \cite{MLDB,bigquery,pmlr-v133-agarwal21a,redshift,postgresql} now support in-database machine learning \cite{10.1145/2213836.2213874,10.1145/3127479.3132746}, where data is stored in relational databases, and machine learning models are trained and executed within database management system. This can improve performance and scalability without extra data transfer overhead. Also, it is natural to express a large class of distributed machine learning (ML) computations relationally. Consider matrix multiplication which is the workhorse of modern ML, assume two matrices \texttt{A} and \texttt{B} which have been partitioned into smaller sub-matrices and stored as relations \cite{luo2018scalable,aggjointree}: 
\begin{SQL}
A(row INT, col INT, mat MATRIX[][])
B(row INT, col INT, mat MATRIX[][])
\end{SQL}

\noindent
A simple SQL code specifies a distributed matrix multiply:

\begin{SQL}
SELECT A.row, B.col,
   SUM (matrix_multiply (A.mat, B.mat))
FROM A, B WHERE A.col = B.row
GROUP BY A.row, B.col
\end{SQL}

In addition to scalability, executing such a code on a relational engine has the advantage that the database query optimizer will automatically distribute the computation, taking into account the sizes of the two matrices.  If \texttt{A} and \texttt{B} are both large matrices, a database optimizer will consider the hardware constraints on each compute node (e.g. memory size) and choose to co-partition both \texttt{A} and \texttt{B} using the join predicate \texttt{A.col = B.row}. If one of the matrices is relatively small (\texttt{A}, for example) and the other matrix is already partitioned across nodes, the database will simply broadcast the smaller matrix. Effectively, the database system is automatically choosing between various distribution paradigms. The first plan is often referred to as \emph{mixed data/model parallelism} or \emph{tensor parallelism} \cite{shazeer2018mesh,MLSYS2019_c74d97b0,shoeybi2019megatron,lepikhin2020gshard,xu2021gspmd,280874,barham2022pathways} in the distributed machine learning literature, and the second plan is \emph{data parallel} \cite{dean2012large,186214}, if \texttt{A} is a model matrix. 

Relational systems can run a wide variety of ML computations \cite{yuan2020tensor,aggjointree}.
For example, consider graph-based convolution operation \cite{GraphConvolutionalNetworks}, which is really a three-way join, followed by an aggregation. Assume data is stored in two relations:

\begin{SQL}
Node(ID INT, vec VECTOR[2048])
Edge(srcID INT, dstID INT)
\end{SQL}

\noindent
Here, \texttt{vec} is the current embedding of a node. Then a graph convolution operation can be written as:

\begin{SQL}
SELECT n1.ID as n.ID, ReLU(MAT_MUL(AVG (Normalize(n2.vec)))) as n.vec
FROM Node as n1, Edge as e, Node as n2
WHERE n1.ID = e.srcID and n2.ID = e.dstID
GROUP BY n1.ID
\end{SQL}

\noindent 
Consider a massive, billion-node, 10-billion-edge graph. Propagating 2048-dimensional embeddings over 10 billion edges will require moving 163 TB of data, which a scalable, distributed database can handle, but will cause problems for most ML systems.

While a relational database may be an excellent platform for executing a large ML computation, ML systems like TensorFlow and PyTorch have at least one key advantage: automatic differentiation \cite{maclaurin2015autograd,seeger2017auto,
van2018automatic,sheldon2018learning,baydin2018automatic,
bolte2020mathematical,
ablin2020super,
lee2020correctness,
oktay2021randomized,
krieken2021storchastic,
bolte2022automatic,
arya2022automatic,
ament2022scalable}.
We argue that without adding an auto-diff capability to relational engines, such compute platforms are unlikely to capture much market share in ML, even for applications to which they are uniquely suited \cite{woznica2005kernels,Moseley2020RelationalAF,pmlr-v139-jayaram21a,sahni2021rasl,pmlr-v133-agarwal21a}.

In this paper, we describe an auto-diff  framework that takes a computation specified in relational algebra (RA) as input and automatically produces a second RA computation that evaluates the gradient of input computation, taken with respect to one or more database tables. A standard SQL compiler and optimizer can further optimize the generated auto-diff'ed SQL programs.
There are several specific contributions of this work:
\begin{itemize}
    \item The gradient operation $\nabla$ is a function-to-function transformation: it takes as input a function $F$, and returns a new function $\nabla F$ that returns the direction of the fastest increase in $F$ from location $\textbf{x}$. Classically, RA is defined operationally (each RA operation takes one or more relations as input and returns a relation). As such, we define a functional version of the RA, as well as gradients of RA functions.
    \item We propose an algorithm for automatically generating a functional RA expression that evaluates the gradient of an input functional RA expression.
    \item We use our RA auto-diff algorithm to automatically produce distributed ML computations and show experimentally that computations generated by RA auto-diff algorithm can easily handle very large-scale ML tasks.
\end{itemize}

\begin{figure}[t]
  \centering
  \begin{tikzpicture}
    \draw \boundellipse{-1.5,0}{0.8}{1.0};
    \draw \boundellipse{1.5,0}{0.8}{1.2};
    \node at (-1.5,1.5) {$K$};
    \node at (1.5,1.5) {$\mathbb{R}^{2\times2}$};
    \node (x1) at (-1.5,0.6) {$\langle 0,0 \rangle$};
    \node (x2) at (-1.5,0.2) {$\langle0,1 \rangle$};
    \node (x3) at (-1.5,-0.2) {$\langle 1,0 \rangle$};
    \node (x4) at (-1.5,-0.6) {$\langle 1,1 \rangle$};
    \node (y1) at (1.5,0.75) {
      $\left[\begin{smallmatrix}
      1 & 4 \\
      1 & 2 \\
    \end{smallmatrix}
    \right]
    $};
    \node (y2) at (1.5,0.25) {
    $\left[\begin{smallmatrix}
      1 & 2 \\
      4 & 3 \\
    \end{smallmatrix}
    \right]
    $};
    \node (y3) at (1.5,-0.25) {
    $\left[\begin{smallmatrix}
      3 & 1 \\
      2 & 2 \\
    \end{smallmatrix}
    \right]
    $};
    \node (y4) at (1.5,-0.75) {
    $\left[\begin{smallmatrix}
      2 & 1 \\
      2 & 2 \\
    \end{smallmatrix}
    \right]
    $};
    \draw[->] (x1) -- (y1);
    \draw[->] (x2) -- (y2);
    \draw[->] (x3) -- (y3);
    \draw[->] (x4) -- (y4);  
  \end{tikzpicture}
  \caption{ $\texttt{R}_\texttt{X}$ as a function from a key set $\{0, 1\} \times \{0, 1\}$ to $\mathbb{R}^{2\times2}$.}
  \vspace{-10 pt}
  \label{fig:relationdiagram1}
\end{figure}

\noindent
\textbf{Roadmap.}
All modern relational database engines are RA engines, executing relational operations (joins, aggregations, and so on) over relations (tables).  Even if a relational computation is expressed in another language (such as SQL), it is compiled into RA.
Thus, in Section \ref{sec:RA}, we define a functional version of RA that is amenable to auto-diff.  In Section \ref{sec:RAAuto-Diff} we re-define partial derivatives, Jacobians, gradients, and vector-Jacobian products in the relational domain.  In Section \ref{sec:RJPs}, we define efficient, relational implementation of  relation-Jacobian products for table scan, selection, aggregation, and join.
Finally, in Section \ref{sec:alg}, we give a relational version of the reverse-mode auto-diff algorithm.

\section{A Functional Relational Algebra} \label{sec:RA}

\subsection{Relations}
Consider a mathematical computation over a set of binary relations $\texttt{R}_0,\texttt{R}_1,\texttt{R}_2,...$. Each relation $\texttt{R}_i$ contains tuples:
$$\texttt{(key, value)}$$
and is a function from some key set $K$ to a value set $V$.  The corresponding function is defined for every $\texttt{key} \in K$. In a sparse representation, a tuple of the form \texttt{(key, value)} may not be present in the underlying implementation of $\texttt{R}$ for each $\texttt{key} \in K$; in such a case, the $\texttt{value}$ associated with a missing $\texttt{key}$ is zero or its equivalent.

We make no assumptions about the form of the key; it may be complex, itself consisting of multiple attributes.  In the general case, $V$ corresponds to all multi-dimensional arrays whose shape is defined by the vector $\textbf{n}$, so $V = \mathbb{R}^{\textbf{n}_1 \times \textbf{n}_2 \times ...}$.

Viewed in this way, relations can easily represent the standard data structures in linear algebra (vectors, matrices, and higher-dimensional tensors) by decomposing the original data structure into tuples holding ``chunks'' or ``blocks''. 
For example:
$$\texttt{X} = \begin{bmatrix}
  1 & 4 & 1 & 2\\
  1 & 2 & 4 & 3\\
  3 & 1 & 2 & 1\\
  2 & 2 & 2 & 2\\
\end{bmatrix}$$
can be decomposed into a relation 
$$\texttt{R}_\texttt{X} (\langle \texttt{rowID}, \texttt{colID} \rangle, \texttt{value})$$ as depicted in Figure \ref{fig:relationdiagram1}.

In the remainder of this section, we make the simplifying assumption that $V = \mathbb{R}$ (so values are all scalars), and define $\allFuncs{K}$ to be the set of all functions from key set $K$ to $\mathbb{R}$; hence, each relation \texttt{R} with key set $K$ is in $\allFuncs{K}$.
However, due to performance considerations, large-scale ML computations implemented relationally should typically be implemented using chunks rather than scalars \cite{luo2018scalable}. Performing computations on a relational engine over a relation storing sub-matrices will give much better performance than over a relation storing a massive number of scalars.  Fortunately, it is straightforward to extend to arrays stored in relations, rather than scalars (as we discuss in Section \ref{sec:extensions}).  

\vspace{5 pt}
\noindent
\textbf{Constant relations vs. queries}.  $\texttt{R}_\texttt{X}$ is an example of a \emph{relation} or a simple function; given any value in the key set, a corresponding real number (or a multi-dimensional tensor in the general case) is produced. However, to build queries, we also need some notion of a higher-order function over relations: a function from one or more input relations to a output relation. It is over such higher-order functions that the gradient operation will itself operate.
Thus, we introduce a \emph{query}, which, given $n$ input key sets $K_1, ..., K_n$ and output key set $K_o$, is a function from relations to a relation: 
\begin{align} 
Q:\allFuncs{K_1}\times \allFuncs{K_2}\times...\times\allFuncs{K_n} \rightarrow \allFuncs{K_o} \nonumber
\end{align}
In the remainder of the paper, we use the simplified notation:
\begin{align} 
Q:\allFuncs{K_1, K_2, ..., K_n} \rightarrow \allFuncs{K_o} \nonumber
\end{align}
Thus, all queries are higher-order functions. A query represents a ``variable relation'' because while the key set is fixed at $K_o$, the actual value from $\allFuncs{K_o}$ taken by the relation depends upon $n$ different input relations given as arguments. 

\subsection{Operations in RA}

Operations such as joins and aggregations in our variant of RA are higher-order functions used to build up queries. We now go through each of the operations, defining each.

\vspace{5 pt}
\noindent 
(1) \emph{TableScan} (denoted using  ``$\tableScan$'') is a higher-order function that accepts a key set $K$ and returns a query that itself accepts a relation in $\allFuncs{K}$ and simply returns exactly that relation.  Formally, $\tableScan$ has type signature
    $\tableScan : K \rightarrow \left(\allFuncs{K} \rightarrow \allFuncs{K} \right)$
and is defined as:
$\tableScan (K) \mapsto \left( \left( \texttt{R} \right) \mapsto \texttt{R} \right).$

\noindent 
(2) \textit{Aggregation} (denoted using ``$\Sigma$'') accepts a query, a grouping function $\grouping$, and a commutative, associative kernel function $\commAssoc$ defined over the real numbers and returns a new query that aggregates the result of the original query.  The function $\grouping : K_{i} \rightarrow K_{o}$ accepts a key value and maps it to a new key value; when the result of a query is aggregated, two tuples $t_1$ and $t_2$ are put into the same group if $\grouping (t_1) = \grouping (t_2)$ and then aggregated using $\commAssoc$. More precisely, aggregation has type signature:
\begin{align}
    \big(\left(K_{i} \rightarrow K_{o} \right) \times \left(\mathbb{R} \rightarrow \mathbb{R} \right) 
    \times \left(\allFuncs{K_1, ..., K_n} \rightarrow \allFuncs{K_{i}} \right) \big) \nonumber \\ 
    \rightarrow \big( \allFuncs{K_1, ..., K_n} \rightarrow \allFuncs{K_{o}} \big) \nonumber
\end{align}
And the semantics of aggregation is as follows:
\begin{align}
 \Sigma \left( \grouping, \commAssoc, Q \right) \mapsto 
   \Big( \left(R_1, R_2, ..., R_n \right) \mapsto \nonumber \Big\{\left(\texttt{key}, \texttt{value} \right) \\
   \textrm{ s.t. for } \texttt{key} \in K_{o} \textrm{ and } Q' = Q(R_1, R_2, ..., R_n), \nonumber \\  \texttt{value} = 
   \commAssoc \left\{ Q'(\texttt{keyIn}) \textrm{ s.t. } \grouping \left( \texttt{keyIn} \right) = \texttt{key}  \right\} \Big\} \Big) \nonumber
\end{align}
A constant grouping function (one that always returns the same value \texttt{key}) aggregates the result of query $Q$ down to a single tuple (for example, holding a scalar loss value). 

Imagine that we wish to represent a four-by-four matrix $\texttt{X}$ relationally, and aggregate its contents down to a single two-by-two matrix. Let $K_\texttt{X}$ denote the key set $\{0, 1\} \times \{0, 1\}$.  We can build up a function $F$ that does exactly this as $F \equiv$
\begin{align}
\Sigma ((\texttt{key}) \mapsto \langle \rangle, (\texttt{v1}, \texttt{v2})
\mapsto \texttt{MatAdd}(\texttt{v1}, \texttt{v2}), \tableScan (K_\texttt{X})) \nonumber
\end{align}
The resulting function $F$ can be applied to any relation in $\allFuncs{K_\texttt{X}}$. For example, $F(\{(\langle 0,0 \rangle, \left[\begin{smallmatrix}
  1 & 4 \\
  1 & 2 \\
\end{smallmatrix}
\right]), (\langle 0,1 \rangle, \left[\begin{smallmatrix}
  1 & 2 \\
  4 & 3 \\
\end{smallmatrix}
\right]),\\ (\langle 1,0 \rangle, \left[\begin{smallmatrix}
  3 & 1 \\
  2 & 2 \\
\end{smallmatrix}
\right]), (\langle 1,1 \rangle, \left[\begin{smallmatrix}
  2 & 1 \\
  2 & 2 \\
\end{smallmatrix}
\right])\})$ evaluates to $\{(\langle \rangle, \left[\begin{smallmatrix}
  7 & 8 \\
  9 & 9 \\
\end{smallmatrix}
\right])\}$.

\vspace{5 pt}
\noindent
(3) \textit{Join} (denoted using ``$\Join$'') accepts two queries $Q_l$ and $Q_r$ (with output key sets $K_l$ and $K_r$, respectively) and produces a new query that composes $Q_l$ and $Q_r$ together, with output key set $K_o$.  In addition to the two queries to compose, $\Join$ accepts three functions: (1) a boolean predicate $\joinPred : K_l \times K_r \rightarrow \{\textrm{true}, \textrm{false}\}$ that takes a key from the key set for $Q_l$ and a key from the key set for $Q_r$ and determines if the two keys match; (2) a projection function $\joinProj : K_l \times K_r \rightarrow K_o$ that accepts a key from the key set for $Q_l$ and a key from the key set for $Q_r$ and composes them; and (3) a kernel function $\joinKern : \mathbb{R} \times \mathbb{R} \rightarrow \mathbb{R}$ that accepts two real-valued \texttt{value}s and composes them.

Formally, the type signature for $\Join$ is:
\begin{align}
 \Join : & \big( \left( K_l \times K_r \rightarrow \{\textrm{true}, \textrm{false}\} \right) \nonumber \\
 &\times \left( K_l \times K_r \rightarrow K_o \right) \times \left( \mathbb{R} \times \mathbb{R} \rightarrow \mathbb{R} \right) \nonumber \\
 &\times \left( \allFuncs{K_{l_1}, K_{l_2}, ..., K_{l_n}} \big) \rightarrow \allFuncs{K_l} \right) \nonumber \\
 &\times ( \allFuncs{K_{r_1}, K_{r_2}, ..., K_{r_m}} \big) \rightarrow
 \allFuncs{K_r} ) \big) \nonumber \\
 &\rightarrow \big( \allFuncs{K_{l_1}, ..., K_{l_n}, K_{r_1}, ..., K_{r_m}} \rightarrow \allFuncs{K_o} \big) \nonumber
\end{align}
And we can define the semantics of $\Join$ as follows:
\begin{align}
\Join &(\joinPred, \joinProj, \joinKern, Q_l, Q_r) \mapsto \nonumber \\
&\Big( \left(R_{l_1}, ..., R_{l_n}, R_{r_1}, ..., R_{r_m} \right) \mapsto \nonumber \Big\{\left(\texttt{key}, \texttt{val} \right) \\
   &\textrm{ s.t. for } \texttt{key} \in K_o \textrm{ and } Q_l' = Q_l(R_{l_1}, ..., R_{l_n}) \nonumber \\ 
   &\textrm{ and } Q_r' = Q_r(R_{r_1}, ..., R_{r_m}),  \nonumber \\ 
   &\texttt{val} = \joinKern \left( Q_l'(\texttt{keyL}), Q_r'(\texttt{keyR}) \right) \nonumber \\ 
   &\textrm{ s.t. } \joinPred (\texttt{keyL}, \texttt{keyR}) = \textrm{ true} \nonumber \\   
   &\wedge  \joinProj (\texttt{keyL}, \texttt{keyR}) = \texttt{key}  \Big\} \Big) \nonumber
\end{align}
We are creating a function that executes queries $Q_l$ and $Q_r$, and then creates a new relation by finding tuples of the form $\left(\texttt{key}, \texttt{val} \right)$ where $\texttt{key}$ is created by applying $\joinProj$ to the keys from two tuples, one from each query result, and $\texttt{val}$ is created by applying  $\joinKern$ to the values from the same tuples.

We can now build up computations such as matrix multiplication.  Given the key set $K_\texttt{X}$, let:
\begin{itemize}
     \item $\commAssoc (\texttt{val1}, \texttt{val2}) \mapsto \texttt{MatAdd}(\texttt{val1}, \texttt{val2})$
     \vspace{-5 pt}
     \item $\grouping (\texttt{key}) \mapsto \langle \texttt{key}[0],\texttt{key}[2] \rangle$
    \vspace{-5 pt}
    \item $\joinKern (\texttt{valL}, \texttt{valR}) \mapsto \texttt{MatMul}(\texttt{valL}, \texttt{valR})$
     \vspace{-5 pt}
     \item $\joinPred (\texttt{keyL}, \texttt{keyR}) \mapsto \texttt{keyL}[1] = \texttt{keyR}[0]$
     \vspace{-5 pt}
     \item $\joinProj (\texttt{keyL}, \texttt{keyR}) \\\mapsto \langle \texttt{keyL}[0], \texttt{keyL}[1],  \texttt{keyR}[1] \rangle$
\end{itemize}
If $K = \{0, 1\} \times \{0, 1\}$, then the following RA builds a function that multiplies two decomposed four-by-four matrices:
\begin{align}
    F_\texttt{MatMul} \equiv \Sigma (\grouping, \commAssoc, \Join (\joinPred, \joinProj, \joinKern, \tableScan (K), \tableScan (K))) \nonumber
\end{align}
\noindent
(4) \textit{Join with one constant input} (denoted using ``$\Join_{\textrm{const}}$''). This is similar to the prior operation, except that one of the inputs to the join is a constant relation, as opposed to a query. Generally, we will not perform gradient descent with respect to \emph{all} relations; some relations must be constant.

$\Join_{\textrm{const}}$ accepts the same three functions as $\Join$: $\joinPred$, $\joinProj$, and $\joinKern$, as well as the query and the constant relation to be joined. The semantics are defined as follows:
\begin{align}
\Join_{\textrm{const}} &(\joinPred, \joinProj, \joinKern, Q, R) \mapsto \Big( \left(R_{l_1}, ..., R_{l_n} \right) \nonumber \\
   &\mapsto \Big\{\left(\texttt{key}, \texttt{value} \right) \nonumber \\
   &\textrm{ s.t. for } \texttt{key} \in K_o \textrm{ and } Q' = Q_l(R_{l_1}, ..., R_{l_n}), \nonumber \\ 
   &\texttt{value} = 
   \joinKern \left( Q'(\texttt{keyInL}), R(\texttt{keyInR}) \right) \nonumber \\
   &\textrm{ s.t. } \joinPred (\texttt{keyInL}, \texttt{keyInR}) = \textrm{ true} \nonumber \\
   &\wedge  \joinProj (\texttt{keyInL}, \texttt{keyInR}) = \texttt{key}  \Big\} \Big) \nonumber
\end{align}
\noindent 
(5) \textit{Selection} (denoted using ``$\sigma$'') builds a function that filters tuples from the output of another query $Q$, but more importantly, the resulting function can modify the \texttt{value}s in the tuples. $\sigma$ accepts three functions: (1) a selection predicate $\selectPred : K_i \rightarrow \{\textrm{true}, \textrm{false}\}$ that takes a key from the input key set for $Q$ and accepts or rejects the key; (2) a projection $\selectProj : K_i \rightarrow K_o$ that modifies the key, (3) a kernel function $\selectKern : \mathbb{R} \rightarrow \mathbb{R}$ that can be used to modify the \texttt{value} in a tuple. Given this, the type signature for $\sigma$ is:
\begin{align}
\sigma : &\big( \left( K_i \rightarrow \{\textrm{true}, \textrm{false}\} \right) \times \left( K_i \rightarrow K_o \right) \times \left( \mathbb{R} \rightarrow \mathbb{R} \right) \nonumber \\ 
&\times \left( \allFuncs{K_{1}, K_{2}, ..., K_{n}} \rightarrow \allFuncs{K_i} \right) \big) \nonumber \\ 
&\rightarrow \left( \allFuncs{K_{1}, K_{2}, ..., K_{n}}  \rightarrow \allFuncs{K_o} \right) \nonumber
\end{align}
And the semantics for $\sigma$ is:
\begin{align}
 \sigma &(\selectPred, \selectProj, \selectKern, Q ) 
 \nonumber \\
&\mapsto \Big( \left(R_{1}, ..., R_{n} \right) \mapsto
   \Big\{\left(\selectProj(\texttt{key}), \texttt{value} \right) \nonumber \\ 
   &\textrm{ s.t. } \selectPred (\texttt{key}) = \textrm{ true}) \nonumber \\
   &\wedge \texttt{for } Q' = Q(R_{1}, ..., R_{n}), \texttt{value} = 
   \selectKern \left(Q'(\textrm{key}) \right)\Big\} \Big) \nonumber
\end{align}

\subsection{Example: Logistic Regression}\label{sec:LR}
For a simple application, we can easily implement logistic regression with cross-entropy loss. Consider the sets: $\texttt{rowID} = \{0, 1, ..., n-1\}$ and $\texttt{colID} = \{0, 1, ..., m-1\}$.  That is, we have $n$ feature vectors identified by the numbers in \texttt{rowID}, each of which has $m$ features identified by the numbers in \texttt{colID}.  Now consider the training set, which consists of feature values for each data point, stored in the relation $\texttt{R}_\texttt{x} \in \allFuncs{\texttt{rowID} \times \texttt{colID}}$, and the set of labels, stored in the relation $\texttt{R}_\texttt{y} \in \allFuncs{\texttt{rowID}}$.   The goal is to learn the set of regression coefficients, stored in the relation $\texttt{R}_\Theta \in \allFuncs{\texttt{colID}}$.  Then the forward pass is:
\begin{align}
F_\texttt{MatMul} \equiv &\Sigma (\grouping_\texttt{MatMul}, \commAssoc, \Join_{\textrm{const}}(
\joinPred_\texttt{MatMul}, \nonumber \\ &\joinProj_\texttt{MatMul}, \joinKern_\texttt{MatMul}, \texttt{R}_\texttt{x},\tableScan (\texttt{colID})) \nonumber \\
F_\texttt{Predict} \equiv &\sigma (\selectPred_\texttt{Predict}, \selectProj_\texttt{Predict}, \selectKern, F_\texttt{MatMul}) \nonumber \\
F_\texttt{Loss} \equiv &\Sigma (\grouping_\texttt{Loss}, \commAssoc, \Join_{\textrm{const}}(\joinPred_\texttt{Loss}, \joinProj_\texttt{Loss}, \nonumber \\ & \joinKern_\texttt{Loss}, F_\texttt{Predict}, \texttt{R}_\texttt{y})) \nonumber
\end{align}
\noindent The matrix multiplication uses functions:
\begin{itemize}
     \item $\commAssoc (\texttt{val1}, \texttt{val2}) \mapsto \texttt{val1} + \texttt{val2}$
     \vspace{-5 pt}
     \item $\grouping_\texttt{MatMul} (\texttt{key}) \mapsto \langle \texttt{key}[0] \rangle$
     \vspace{-5 pt}
     \item $\joinKern_\texttt{MatMul} (\texttt{valL}, \texttt{valR}) \mapsto \texttt{valL} \times \texttt{valR}$
     \vspace{-5 pt}
     \item $\joinPred_\texttt{MatMul} (\texttt{keyL}, \texttt{keyR}) \mapsto \texttt{keyL}[1] = \texttt{keyR}[0]$
     \vspace{-5 pt}
     \item $\joinProj_\texttt{MatMul} (\texttt{keyL}, \texttt{keyR}) \mapsto \langle \texttt{keyL}[0], \texttt{keyL}[1] \rangle$
\end{itemize}
The selection utilizes a logistic function to make predictions:
\begin{itemize}
     \item $\selectPred_\texttt{Predict} (\texttt{key}) \mapsto \texttt{true}$
     \vspace{-5 pt}
     \item $\selectProj_\texttt{Predict} (\texttt{key}) \mapsto \texttt{key}$
     \vspace{-5 pt}
     \item $\selectKern (\texttt{val}) \mapsto \texttt{logistic(val)}$
\end{itemize}
And a cross-entropy loss computes the quality of the model:
\begin{itemize}
     \item $\grouping_\texttt{loss} (\texttt{key}) \mapsto \langle \rangle$
     \vspace{-5 pt}
     \item $\joinKern_\texttt{Loss} (\texttt{yhat}, \texttt{y}) \mapsto -\texttt{y} \log \texttt{yhat} + (\texttt{y} - 1) \log (1 - \texttt{yhat}) $
     \vspace{-5 pt}
     \item $\joinPred_\texttt{Loss} (\texttt{keyL}, \texttt{keyR}) \mapsto \texttt{keyL}[0] = \texttt{keyR}[0]$
     \vspace{-5 pt}
     \item $\joinProj_\texttt{Loss} (\texttt{keyL}, \texttt{keyR}) \mapsto \langle \texttt{keyL}[0] \rangle$
\end{itemize} 
Now, $F_\texttt{Loss}$ is a function from $\allFuncs{\texttt{colID}}$ to $\allFuncs{ \{\langle \rangle\}}$.  That is, executing the query $F_\texttt{Loss}$ on a relation that contains all of the regression coefficients will return a simple tuple with empty key $\langle \rangle$ and whose value contains the loss. Figure \ref{fig:autodiff} (in appendix) shows this example on the left part. 

\section{Auto-Diffing RA: Preliminaries}
\label{sec:RAAuto-Diff}

\subsection{Relational Partial Derivatives and Jacobians}

Our goal is ultimately to perform auto-differentiation on functions such as $F_\texttt{Loss}$ to power standard optimization algorithms such as gradient descent.  To do this it is first necessary to re-define standard concepts such as partial-derivatives and Jacobians in the relational domain.

\vspace{5 pt}
\noindent
\textbf{Relational partial derivatives.}
Consider any query $Q : \allFuncs{K_i} \rightarrow \allFuncs{K_o}$.\footnote{In the case that $Q$ takes $n > 1$ arguments, all of the definitions in this section apply to $Q$, given $n - 1$ constant relations and partially applying $Q$ to those $n - 1$ relations to obtain a one-argument function.} We denote the \emph{partial derivative of} $Q$ \emph{with respect to} a tuple $(k,v)$ with $k \in K_i$ by $\frac{\partial Q}{\partial k}$. This partial derivative is itself a function with type signature $\allFuncs{K_i} \rightarrow \allFuncs{K_o}$. 

To formally define this concept---which is  analogous to the partial derivative of a multi-variate function $F$ with respect to a particular input, consider the relation $\texttt{R}_h$ in $\allFuncs{K_i}$ where $\texttt{R}_h[k]$ is $h$, and $\texttt{R}_h[k']$ is $0$ for $k' \neq k$. Now, let:
\begin{itemize}
    \item $\joinKern_1 (\texttt{valL}, \texttt{valR}) \mapsto \texttt{valL} + \texttt{valR}$
     \vspace{-5 pt}
        \item $\joinKern_2 (\texttt{valL}, \texttt{valR}) \mapsto \frac{\texttt{valR} - \texttt{valL}}{h}$
     \vspace{-5 pt}
     \item $\joinPred (\texttt{keyL}, \texttt{keyR}) \mapsto \texttt{keyL} = \texttt{keyR}$
     \vspace{-5 pt}
     \item $\joinProj (\texttt{keyL}, \texttt{keyR}) \mapsto \texttt{keyL}$
\end{itemize}
Now, we can define $\frac{\partial Q}{\partial k}$ as the function:
\begin{align}
(\texttt{R}) &\mapsto \lim_{h \rightarrow 0} 
\Join\Big(\joinPred, \joinProj, \joinKern_2,Q,\nonumber \\
&Q\big(\Join_{\textrm{const}} (\joinPred, \joinProj, \joinKern_1, \texttt{R}_h, \tableScan (K_i))\big)\Big) (\texttt{R}, \texttt{R}) \nonumber
\end{align}
This is the query that we obtain by creating a ``slightly'' perturbed version of $Q$ that adds $h$ to the value associated with key $k$. We run $Q$ on input relation \texttt{R} as well as the perturbed version of $Q$ on \texttt{R}, and then join the output of the two versions of $Q$ to compute how much each output tuple varies.

\vspace{5 pt}
\noindent
\textbf{Relational Jacobians.} In real analysis, a \emph{Jacobian} is a matrix of functions, where each function is the partial derivative of a multivariate function with respect to a unique input/output variable pair. We denote a Jacobian of query $Q$ as $J_{Q} : \allFuncs{K_i} \rightarrow \allFuncs {K_i \times K_o}$. Consider the functions:
\begin{itemize}
   \item $\selectPred (\texttt{key}) \mapsto \texttt{key}[0] = k$
   \vspace{-5 pt}
   \item $\selectProj (\texttt{key}) \mapsto \texttt{key}[1]$
   \vspace{-5 pt}
   \item $\selectKern (\texttt{val}) \mapsto \texttt{val}$
\end{itemize}
$J_Q$ is the Jacobian for query $Q$, if for every key $k \in K_i$:
$$\sigma(\selectPred, \selectProj, \selectKern, J_Q) \equiv \frac{\partial Q}{\partial k}.$$
That is, the relational Jacobian is a query that performs a relational partial derivative for every possible input key.

\vspace{5 pt}
\noindent
\textbf{Relational gradients.}
Define the \emph{gradient of query} $Q : \allFuncs{K_i} \rightarrow \allFuncs{K_o}$ with respect to $k \in K_o$ in terms of the Jacobian.
Let:  $\selectPred (\texttt{key}) \mapsto \texttt{key}[1] = k$, $\selectProj (\texttt{key}) \mapsto \texttt{key}[0]$, $\selectKern (\texttt{val}) \mapsto \texttt{val}$.
Then the gradient of query $Q$  with respect to key $k$ is:
$$\nabla_k Q \equiv \sigma(\selectPred, \selectProj, \selectKern, J_Q)$$
To obtain the gradient, we restrict the Jacobian of query $Q$ to one of the keys in the output set, filtering out the rest.  Note that if $Q$ has only one output tuple---if it is computing a loss value, for example---then the Jacobian of $Q$ and the gradient of $Q$ are essentially equivalent, in the sense that evaluating either over a relation $\texttt{R} \in \allFuncs{K_i}$ will produce singleton relations with tuples having the same \texttt{value}s.  In this case, we drop the key $k$ and write $\nabla Q$.

\vspace{5 pt}
\noindent
\textbf{Multi-relation queries.} In the case where a query $Q$ has multiple table scans (and hence takes multiple relations as inputs), the notions of relational Jacobian and relational gradients still apply.  These are defined by picking the table scan $\tableScan_i$ associated with the $i$th input relation, and partially evaluating $Q$ using given, constant relations for each table scan $\tableScan \neq \tableScan_i$.  This results in a single-argument query, which we refer to using $Q_i$.  The relational Jacobian and relational gradients are then defined with respect to $Q_i$.  For an ML computation encoded as a relational query with $n$ input relations (whose values we want to learn via some form of gradient descent) having current values $\texttt{R}_1, \texttt{R}_2, ..., \texttt{R}_n$, we would typically want to evaluate $\nabla Q_1 (\texttt{R}_1), \nabla Q_2 (\texttt{R}_2), ..., \nabla Q_n (\texttt{R}_n)$ to power gradient descent.  This is the topic we consider in the next few sections of the paper.

\subsection{Relation-Jacobian Products}
\label{sec:RJP}

As our goal is to build a reverse-mode, relational auto-diff engine, we next define the analog to the vector-Jacobian product in the relational domain, which we call the \emph{relation-Jacobian product}. Assume that we have a query $Q : \allFuncs{K_i} \rightarrow \allFuncs{K_o}$.  Let:
\begin{itemize}
     \item $\commAssoc (\texttt{val1}, \texttt{val2}) \mapsto \texttt{val1} + \texttt{val2}$
     \vspace{-5 pt}
     \item $\grouping (\texttt{key}) \mapsto \texttt{key}[0]$
     \vspace{-5 pt}
    \item $\joinKern (\texttt{valL}, \texttt{valR}) \mapsto \texttt{valL} \times \texttt{valR}$
    \vspace{-5 pt}
     \item $\joinPred (\texttt{keyL}, \texttt{keyR}) \mapsto \texttt{keyL} = \texttt{keyR}[1]$
     \vspace{-5 pt}
     \item $\joinProj (\texttt{keyL}, \texttt{keyR}) \mapsto   \texttt{keyR}$
\end{itemize}
Then the \emph{relation-Jacobian product for query} $Q$, denoted $RJP_Q : \allFuncs{K_o, K_i} \rightarrow \allFuncs{K_i}$ is defined as:
\begin{align}
RJP_Q \equiv \Sigma (\grouping, \commAssoc, \Join (\joinPred, \joinProj, \joinKern, \tableScan (K_o), J_Q)) \nonumber
\end{align}

\begin{algorithm}[t]
  \caption{ChainRule ($v_i$, $v_j$, $\frac{\partial Q}{\partial \texttt{R}_j}$, $\langle \texttt{R}_1, ..., \texttt{R}_k \rangle$)}
  \label{alg:chain}
  \begin{algorithmic}[1]
    \STATE \textbf{Input}: Connected RA operations $v_i$, $v_j$ (the result of $v_i$ is input into $v_j$), relation $\frac{\partial Q}{\partial \texttt{R}_j}$, list of all intermediate results obtained when executing $Q$: $\langle \texttt{R}_1, ..., \texttt{R}_k \rangle$ 
    \STATE \textbf{Output}: $\frac{\partial Q}{\partial \texttt{R}_i}$ computed via the chain rule
    \STATE Let $K(v)$ denote the output key set for  RA operation $v$
    \IF{$v_j$ is $\Sigma \left( \grouping, \commAssoc, v_i \right)$} 
        \STATE $\frac{\partial Q}{\partial \texttt{R}_i} \leftarrow RJP_{\Sigma}(\grouping, \commAssoc, K(v_j), K(v_i)) (\frac{\partial Q}{\partial \texttt{R}_j}, \texttt{R}_i)$
    \ELSIF{$v_j$ is $\sigma \left( \selectPred, \selectProj, \selectKern, v_i \right)$}
        \STATE $\frac{\partial Q}{\partial \texttt{R}_i} \leftarrow RJP_{\sigma}(\selectPred, \selectProj, \selectKern, \newline K(v_j), K(v_i)) (\frac{\partial Q}{\partial \texttt{R}_j}, \texttt{R}_i)$
    \ELSIF{$v_j$ is $\tableScan \left( K \right)$}
        \STATE $\frac{\partial Q}{\partial \texttt{R}_i} \leftarrow RJP_{\tableScan}(K) (\frac{\partial Q}{\partial \texttt{R}_j}, \texttt{R}_i)$
    \ELSIF{$v_j$ is $\Join (\joinPred, \joinProj, \joinKern, v_i, v_k)$}
        \STATE $\frac{\partial Q}{\partial \texttt{R}_i} \leftarrow RJP_{\Join}(\joinPred, \joinProj, \joinKern, \newline K(v_j), K(v_i), \texttt{R}_k)(\frac{\partial Q}{\partial \texttt{R}_j}, \texttt{R}_i)$
    \ELSIF{$v_j$ is $\Join (\joinPred, \joinProj, \joinKern, v_k, v_i)$}
        \STATE $\frac{\partial Q}{\partial \texttt{R}_i} \leftarrow RJP_{\Join}(\joinPred, \joinProj, \joinKern, \newline K(v_j), \texttt{R}_k, K(v_i)) (\frac{\partial Q}{\partial \texttt{R}_j}, \texttt{R}_i)$
    \ELSIF{$v_j$ is $\Join_{\texttt{const}} (\joinPred, \joinProj, \joinKern, v_i, \texttt{R})$}
        \STATE $\frac{\partial Q}{\partial \texttt{R}_i} \leftarrow RJP_{\Join}(\joinPred, \joinProj, \joinKern, \newline K(v_j), K(v_i), \texttt{R}) (\frac{\partial Q}{\partial \texttt{R}_j}, \texttt{R}_i)$
    \ENDIF
    \STATE Return $\frac{\partial Q}{\partial \texttt{R}_i}$
  \end{algorithmic}
\end{algorithm}

\begin{algorithm}[t]
  \caption{RAAutoDiff ($Q$, $\langle \texttt{In}_1, \texttt{In}_2, ... \rangle $)}
  \label{alg:auto}
  \begin{algorithmic}[1]
    \STATE \textbf{Input}: Query $Q$ computing a one-tuple loss, list of input relations $\langle \texttt{In}_1, \texttt{In}_2, ... \rangle$
    \STATE \textbf{Output}: $\nabla Q_1 (\texttt{In}_1), \nabla Q_2 (\texttt{In}_2), ...$
    \STATE Topologically sort RA operations in $Q$ into a list of operations  $\langle v_{1},\ldots,v_{n}\rangle$
    \STATE Let $E$ be the list of edges in $Q$, where $(v_i, v_j) \in E$ if the output of $v_i$ us used by $v_j$
    \STATE Execute $Q (\texttt{In}_1, \texttt{In}_2, ...)$
    \STATE Let $\texttt{R}_i$ denote the intermediate relation produced by $V_i$ for each $i \in \{1...n\}$ during execution
    \STATE Set $\frac{\partial Q}{\partial \texttt{R}_n}$ to $\{(\texttt{keyOut}, 1)\}$ where \texttt{keyOut} is the key in the output tuple from $Q$
    \FOR{$i = n - 1$ down to $1$}
        \STATE \% Compute $\frac{\partial Q}{\partial \texttt{R}_i}$
        \STATE Let $I = \langle id_1, id_2, .., id_m \rangle $ be a list of vertex identifiers s.t. $id_j \in I$ if $(v_i, v_{id_j}) \in  V$
        \STATE $\texttt{P}_1$ $\leftarrow$ ChainRule ($v_i$, $v_{id_1}$, $\frac{\partial Q}{\partial \texttt{R}_{id_1}}$, $\langle \texttt{R}_1, ..., \texttt{R}_k \rangle$)
        \STATE Let $K$ be the key set for $\texttt{R}_i$
        \STATE $Q' \leftarrow \tableScan (K)$
        \FOR{$j = 2$ to $m$}
            \STATE $\texttt{P}_j$ $\leftarrow$ ChainRule ($v_i$, $v_{id_j}$, $\frac{\partial Q}{\partial \texttt{R}_i}$, $\langle \texttt{R}_1, ..., \texttt{R}_k \rangle$)
            \STATE $Q' \leftarrow \add (Q', \tableScan (K))$
        \ENDFOR
        \STATE $\frac{\partial Q}{\partial \texttt{R}_i} \leftarrow Q' (\texttt{P}_1, \texttt{P}_2, ..., \texttt{P}_m)$
    \ENDFOR
    \STATE For the $i$th input to $Q$, find the $v_j$ that processed $\texttt{In}_i$ as input.  Return the associated $\frac{\partial Q}{\partial \texttt{R}_j}$ as $\nabla Q_i(\texttt{In}_i)$.
  \end{algorithmic}
\end{algorithm}

\section{RJPs for Relational Operations}\label{sec:RJPs}
Many auto-diff engines work by first executing the underlying computation, collecting intermediate results, during a forward pass.  Then those results are used to evaluate the desired gradient(s) in a backward pass, via a series of vector-Jacobian products.  

Thus, there are two key parts of any classical reverse-mode auto-diff system: (1) the overall algorithmic framework that runs the forward and backward passes, and (2) vector-Jacobian product implementations for each RA operation.

To support auto-diff for RA, we need something analogous to both of these parts.  In this section of the paper, we describe relation-Jacobian product (RJP) implementations for each of the higher-order RA functions we have defined.  

\vspace{5 pt}
\noindent
\textbf{RJP for Table Scan.}
Consider query $Q \equiv \tableScan(K)$, for a key set $K$.  The RJP for this query, denoted as $RJP_{\tableScan}(K) : \allFuncs{K, K} \rightarrow \allFuncs{K}$, can be computed as:
$$RJP_{\tableScan}(K) \mapsto ((\texttt{R}_o, \texttt{R}_i) \mapsto \texttt{R}_o)$$
This RJP is simple because the table scan returns its input relation; $\left(J_Q\left(\texttt{R}_i\right)\right)(\langle k_1, k_2 \rangle)$ then is one for any $k_1, k_2 \in K$ where $k_1 = k_2$, and zero when $k_1 \neq k_2$; taking the left product with $\texttt{R}_o$ as defined in Section \ref{sec:RJP} simply returns $\texttt{R}_o$, no matter the value of $\texttt{R}_i$.

\vspace{5 pt}
\noindent
\textbf{RJP for Selection.}
Consider the query $Q \equiv \sigma (\selectPred, \selectProj, \selectKern, \tableScan(K_i))$, with type signature $Q : \allFuncs{K_i} \rightarrow \allFuncs{K_o}$.  The RJP for $Q$, $RJP_{\sigma}(\selectPred, \selectProj, \selectKern, K_o, K_i) : \allFuncs{K_o, K_i} \rightarrow \allFuncs{K_i}$, is: 
\begin{align}
\Join (\joinPred', \joinProj', \joinKern', \tableScan(K_o), \tableScan(K_i)) \nonumber
\end{align}
where:
\begin{itemize}
    \item $\joinPred' (\texttt{keyL}, \texttt{keyR}) \mapsto \texttt{keyL} = \selectProj(\texttt{keyR})$
    \vspace{-5 pt}
    \item $\joinProj' (\texttt{keyL}, \texttt{keyR}) \mapsto \texttt{keyR}$
    \vspace{-5 pt}
    \item $\joinKern' (\texttt{valL}, \texttt{valR}) \mapsto 
    \frac{\partial \selectKern(\texttt{valR})}{\partial \texttt{valR}} \times \texttt{valL} $
\end{itemize}
Here, $
    \frac{\partial \selectKern(\texttt{valR})}{\partial \texttt{valR}}$ is the derivative of $\selectKern\texttt{(valR)}$ w.r.t. input $\texttt{valR}$.
    Note that $\sigma$ will discard some tuples if they cannot meet the boolean condition specified in $\selectPred$. Those tuples tuples cannot contribute to a gradient computation, and the gradient evaluated at a key value that has been filtered from the relation will implicitly be zero.

\vspace{5 pt}
\noindent
\textbf{RJP for Aggregation.} Consider the query $Q \equiv \Sigma (\grouping, \commAssoc, \tableScan(K_i))$, with type signature $Q : \allFuncs{K_i} \rightarrow \allFuncs{K_o}$.  The RJP for this query,  denoted as $RJP_{\Sigma}(\grouping,\commAssoc, K_o, K_i) : \allFuncs{K_o, K_i} \rightarrow \allFuncs{K_i}$ is:
\begin{align}
\Join (\joinPred, \joinProj, \joinKern, \tableScan(K_o), \tableScan(K_i)) \nonumber
\end{align}
where: $\joinPred (\texttt{keyL}, \texttt{keyR}) \mapsto \texttt{keyL} = \grouping(\texttt{keyR})$,  $\joinProj (\texttt{keyL}, \texttt{keyR}) \mapsto \texttt{keyR}$, $\joinKern (\texttt{valL}, \texttt{valR}) \mapsto 
    \frac{\partial \commAssoc(\texttt{valR})}{\partial \texttt{valR}} \times \texttt{valL}$.
Here, $\frac{\partial \commAssoc(\texttt{valR})}{\partial \texttt{valR}}$ is the derivative function of $\commAssoc\texttt{(valR)}$ w.r.t. input $\texttt{valR}$.
If $\grouping$ is a constant function, the RJP can be simplified to:
\begin{align}
&RJP_{\Sigma}(\grouping,\commAssoc, K_o, K_i) \nonumber \\
&\mapsto ((\texttt{R}_o, \texttt{R}_i) \mapsto \sigma(\selectPred, \selectProj, \selectKern,\tableScan(K_i))(\texttt{R}_i)) \nonumber
\end{align}
where: $\selectPred (\texttt{key}) \mapsto \texttt{true}$, $\selectProj (\texttt{key}) \mapsto \texttt{key}$, $\selectKern (\texttt{val}) \mapsto
   \frac{\partial \commAssoc(\texttt{val})}{\partial \texttt{val}}$.

\vspace{5 pt}
\noindent
\textbf{RJP for Join.} Consider the query $Q$ that computes $$\Join (\joinPred, \joinProj, \joinKern, \tableScan(K_l),\tableScan(K_r))$$ 
with type signature $Q : \allFuncs{K_l,K_r} \rightarrow \allFuncs{K_o}$. Since this query has two inputs, we first consider computing the RJP for query $Q_l$. That can be obtained by partially evaluating $Q$ with the constant relation $\texttt{R}_r$, so that $Q_l : \allFuncs{K_l} \rightarrow \allFuncs{K_o}$.
\begin{align*}
RJP_{\Join}&(\joinPred, \joinProj, \joinKern, K_o, K_l, \texttt{R}_r) \equiv \\
\Sigma &(\grouping, \commAssoc(\Join (\joinPred_1, \joinProj_1, \joinKern_1, \tableScan({K}_o), \\
&\Join_{\textrm{const}} (\joinPred_2, \joinProj_2, \joinKern_2, \tableScan(K_l), \texttt{R}_r))))
\end{align*}
where:
\begin{itemize}
   \item $\grouping(\texttt{key}) \mapsto \langle \texttt{key} \rangle$
   \vspace{-5pt}
   \item $\commAssoc(\texttt{v1},\texttt{v2}) \mapsto \texttt{v1} + \texttt{v2}$
   \vspace{-5pt}
   \item $\joinPred_1(\texttt{keyL}, \texttt{keyR}) \mapsto \texttt{keyL} = \texttt{keyR}[1]$
   \vspace{-5pt}
   \item $\joinProj_1(\texttt{keyL}, \texttt{keyR}) \mapsto \texttt{keyR}[0]$
   \vspace{-5pt}
   \item $\joinKern_1(\texttt{valL}, \texttt{valR}) \mapsto \texttt{valL} \times \texttt{valR}$
   \vspace{-5pt}
   \item $\joinPred_2(\texttt{keyL}, \texttt{keyR}) \mapsto \joinPred(\texttt{keyL}, \texttt{keyR})$
   \vspace{-5pt}
   \item $\joinProj_2(\texttt{keyL}, \texttt{keyR}) \\\mapsto \langle \texttt{keyL}, \joinProj(\texttt{keyL}, \texttt{keyR}) \rangle$
   \vspace{-5pt}
   \item $\joinKern_2(\texttt{valL}, \texttt{valR}) \mapsto \frac{\partial \joinKern(\texttt{valL}, \texttt{valR})}{\partial \texttt{valL}}$
\end{itemize}
If the query $Q \equiv \textrm{} \Join_{\textrm{const}} (\joinPred, \joinProj, \joinKern, \tableScan(K_l), \texttt{R}_r)$ so that the right-hand relation is a constant, then the RJP is exactly the same; the RJP for this query is also $RJP_{\Join}(\joinPred, \joinProj, \joinKern, K_o, K_l, \texttt{R}_r)$. If the goal is to compute the RJP of $Q_r$ (that is, where the left-hand relation is constant) things are symmetric and defined similarly, we denote the RJP in this case using $RJP_{\Join}(\joinPred, \joinProj, \joinKern, K_o, \texttt{R}_l, K_r)$.

There are some further optimization opportunities for $RJP_{\Join}$: 
\begin{itemize}
    \item The first $\Join_{\textrm{const}}$ operation can often be optimized out since most ML workloads fix $\joinKern$ to be $\times$ (or $\texttt{MatMul}$). For the RJP of $Q_l$ and $Q_r$, the result of $\Join_{\textrm{const}}$ can be replaced by $\texttt{R}_r$ and $\texttt{R}_l$, respectively.
    \item The final $\Sigma$ operation can be optimized out based on different join cardinality relationships (one-to-one, one-to-many). If $\Join$ is $\Join_{1-1}$, the $\Sigma$ for RJP of $Q_l$ and  $Q_r$ can be directly removed. If $\Join$ is $\Join_{1-n}$ or $\Join_{n-1}$: for the $\texttt{n}$ side, $\Sigma$ can be optimized in the same way, while for the $\texttt{1}$ side, the $\Sigma$ must be kept since each tuple's partial gradients needed to be aggregated. 
    \item When a join-agg-tree structure ~\cite{aggjointree} (a join followed by an aggregation) appears in query graph, differentiating the aggregation operator is unnecessary. 
\end{itemize}

\section{Relational Auto-Differentiation}
\label{sec:alg}
We are now ready to give the final algorithm for relational auto-diff.
To give the formal algorithm, we first define the relational $\add$ operation, that takes two relations $Q_l, Q_r \in \allFuncs{K}$ and is defined as $\add (Q_l, Q_r) \mapsto$
\begin{align}
&\Big( \left(R_{l_1}, ..., R_{l_n}, R_{r_1}, ..., R_{r_m} \right) \mapsto \Big\{\left(\texttt{key}, \texttt{value} \right) \nonumber \\
   &\textrm{ s.t. for } \texttt{key} \in K \textrm{ and }Q_l' = Q_l(R_{l_1}, ..., R_{l_n}) \textrm{ and }  \nonumber \\ 
   &Q_r' = Q_r(R_{r_1}, ..., R_{r_m}), \nonumber \\ 
   &\texttt{value} = Q_l'(\texttt{key}) + Q_r'(\texttt{key})  \Big\} \Big) \nonumber
\end{align}
\texttt{add} takes two queries with the same key set and returns a new query that adds the \texttt{value}s with matching keys across queries.  \texttt{add} is necessary to implement the total derivative.

The final algorithm is given as the subroutine Algorithm \ref{alg:chain} and the main procedure Algorithm \ref{alg:auto}.  Algorithm \ref{alg:chain} implements the chain rule for each of the various relational operations using RJPs.  Algorithm \ref{alg:auto} performs the actual reverse-mode auto-diff, first running the query and then going through the various relational operations in the query in reverse topological order.  For each RA operation, the chain rule is used to compute $\frac{\partial Q}{\partial \texttt{R}_i}$, via the appropriate RJP.  Figure \ref{fig:compare-LA-RA-autodiff} (in appendix) shows the difference between vector-jacobian product and relational-jacobian product for a single matrix multiplication operation. Figure \ref{fig:autodiff} (in appendix) shows Algorithm \ref{alg:auto} in the right part.

\begin{table}[t]
\small\setlength\tabcolsep{3pt}
  \begin{tabular}{|l|l|c|c|}
    \hline
    Dataset name & $(|V|,|E|)$ & Num feat & Num labels\\ \hline
    \texttt{ogbn-arxiv} & (0.2M, 1.1M) & 128 & 40 \\
    \texttt{ogbn-products} & (0.1M, 39M) & 100 & 47 \\
    \texttt{ogbn-papers100M} & (0.1B, 1.6B) & 128 & 172 \\
    \texttt{friendster} & (65.6M, 3.6B) & 128 & 100 \\
    \hline
  \end{tabular}
  \caption{Data used for graph convolutional network training.}
  \vspace{-10 pt}
  \label{table:gcndataset}
\end{table}

\begin{table*}[h]
  \begin{center}
  \small\setlength\tabcolsep{7pt}
  \begin{tabular}{|c|c|c|c|c|c|c|c|c|c|c|}
    \hline
     & \multicolumn{5}{c|}{\texttt{ogbn-arxiv}} & \multicolumn{5}{c|}{\texttt{ogbn-products}} \\
    \hline
    Cluster Size & 1 & 2 & 4 & 8 & 16 & 1 & 2 & 4 & 8 & 16\\
    \hline
    \texttt{DistDGL} &  1.664s  &  1.407s &  0.731s  & 0.483s &  0.321s & 14.827s & 9.270s & 4.980s &  2.889s & 1.799s \\
    \texttt{AliGraph} &  13.734s  &  5.488s  & 3.603s & 1.744s &  1.564s &  87.299s &  55.193s	& 31.128s &  17.303s & 11.734s \\
    \textbf{RA-GCN} &  \textbf{9.957s} & \textbf{5.125s}  & \textbf{2.741s} &  \textbf{1.604s} & \textbf{0.957s} & \textbf{31.347s} & \textbf{16.409s} & \textbf{10.713s} & \textbf{6.873s} & \textbf{4.591s}\\
    \textbf{RA-GCN(full)}  & \textbf{20.196s} & \textbf{11.739s} & \textbf{7.338s} & \textbf{4.331s} & \textbf{3.196s} & \textbf{54.424s} & \textbf{33.185s} & \textbf{19.028s} & \textbf{13.572s} & \textbf{9.897s}\\
    \hline
  \end{tabular}
  \caption{Distributed graph convolutional network runtime per epoch on \texttt{ogbn-arxiv} and \texttt{ogbn-products}. \texttt{RA-GCN} (full) is the experiment results for full graph training while others are mini-batch based training.}
  \label{table:gcnruntimesmall}
  \end{center}
\end{table*}

\begin{table*}[t]
  \begin{center}
    \small\setlength\tabcolsep{4pt}
    \begin{tabular}{|c|c|c|c|c|c|c|c|c|c|c|}
      \hline
       & \multicolumn{5}{c|}{\texttt{ogbn-papers100M}} & \multicolumn{5}{c|}{\texttt{friendster}} \\
      \hline
      Cluster Size & 1 & 2 & 4 & 8 & 16 & 1 & 2 & 4 & 8 & 16\\
      \hline
      \texttt{DistDGL} &  OOM  &  OOM &  71.842s  & 56.517s & 39.824s & OOM & OOM & OOM & 92.741 &  71.826s\\
      \texttt{AliGraph}   &  OOM  &  OOM  & OOM &  OOM & OOM &  OOM &  OOM &  OOM & OOM &  OOM\\
     \textbf{RA-GCN} &  \textbf{295.184s} & \textbf{154.94s}  & \textbf{78.091s} &  \textbf{52.937s} & \textbf{36.409s} & \textbf{371.572s} & \textbf{194.212s} & \textbf{125.405s} & \textbf{87.913s} & \textbf{63.354s}\\
      \textbf{RA-GCN(full)} & \textbf{1161.553s} & \textbf{614.121s} & \textbf{327.609s} & \textbf{218.339s} & \textbf{133.581s} & \textbf{1492.142s} & \textbf{781.102s} & \textbf{485.247s} & \textbf{317.087s} & \textbf{279.763s}\\
      \hline
    \end{tabular}
  \caption{Distributed graph convolutional network runtime per epoch on \texttt{ogbn-papers100M} and \texttt{friendster}.}
  \vspace{-10 pt}
  \label{table:gcnruntimelarge}
  \end{center}
\end{table*}

\section{Evaluation}

One potential benefit of relational auto-diff is that a relational system, equipped with this technology, could show better scalability than other systems.  Hence we turn our attention to the question:
\emph{Can relational auto-diff be used to produce computations that are competitive with special-purpose ML systems meant to support large-scale machine learning?} 

Our evaluation focuses on three distributed ML computations over big data: graph convolutional neural networks \cite{GraphConvolutionalNetworks,liu2022rsc,liu2021exact,zhou2020towards,pipegcn,bns-gcn}, knowledge graph embedding \cite{knowledgegraph}, and non-negative matrix factorization \cite{NIPS2000_f9d11525} (the latter two are relegated to the Appendix). We implemented  RA auto-diff in Python, accepting SQL input.

Experiments are run on AWS, using \texttt{m5.4xlarge} instances with 20 cores, 64GB DDR4 memory, and 1TB general SSD.  We run our experiments using 1 to 16 nodes connected by 10Gbps Ethernet. We build our relational computations on top of a relational engine \cite{plinycompute}. It is worth to mention here all the RA operators, RJP rules, and related implementation in this paper can easily be incorporated into any relational system that supports array types.

\noindent \textbf{Task Evaluated.} We benchmark a two-layer, graph convolutional neural network (GCN) for a node classification task. A graph convolutional layer can be easily written as a relational computation over two relations $\texttt{Edge}$ (storing all the edges including self-loops, each having a normalized weight) and
$\texttt{Node}$ (storing all the node embeddings in the graph).
Message passing across nodes is implemented as a three-way join among nodes, edges, and nodes, followed by an aggregation.
This join extracts the \texttt{ID} from both source node and destination nodes,
and matches them with the \texttt{sourceID} and \texttt{destID} of the edges. This GCN is benchmarked using the datasets in Table ~\ref{table:gcndataset}.

\noindent 
\textbf{Experiments.}
We compare against two other state-of-the-art open-source graph systems: \texttt{DistDGL} ~\cite{distdgl} and \texttt{AliGraph} ~\cite{aligraph}. \texttt{RA-GCN} is our RA-based implementation.  The Adam optimizer is used with learning rate $\eta = 0.1$; the
dropout rate $\gamma = 0.5$; 
the hidden layer dimension $D = 256$; batch size $B = 1024$.
\texttt{DGL} is built from the latest version 0.9 from scratch.
We use PyTorch ~\cite{pytorch} distributed as the backend for \texttt{AliGraph}. All of the systems are running the same learning computations over the same input data, using the same batch size, the same initial data partitioning scheme, and the same model.

As a scalable, RA-based system, \texttt{RA-GCN} is able to handle arbitrary-size batches and can even perform full graph training, 
while the other systems can only support ``data-parallel" graph training, partitioning large graphs into sub-graphs and sampling neighbors to form mini-batches. We also include full-graph training on \texttt{RA-GCN}.
For each of the four datasets and four methods tested, per-epoch running times are shown in Tables ~\ref{table:gcnruntimesmall} and ~\ref{table:gcnruntimelarge}. ``OOM'' denotes the case that a system failed due to out-of-memory errors.

\noindent 
\textbf{Discussion.}
Our experiments generally showed that executing the RA-based auto-diff output consistently results in a computation that is as fast as the state-of-the-art alternatives. The only exceptions were for the GCN runs over the smallest data sets (\texttt{ogbn-arxin} and \texttt{ogbn-products}), where the RA-based solution was somewhat slower than its competitors.  This is perhaps not surprising: one might not expect the benefit of a scalable, RA-based solution to be apparent over a very small data set, compared to a custom-built ML solution. 

However, there were some clear advantages of the auto-diffed RA solution.  As the auto-diffed RA is running on what is essentially a high-performance database system, 
it avoided all out-of-memory errors. \texttt{RA-GCN} was able to scale to the largest data set (\texttt{friendster}), even for full graph training--thus avoiding the potential pitfalls of cutting important edges during training.  In fact, the RA-based solution was the only solution able to scale to full-graph training. Further, \texttt{RA-GCN} was able to do this on only one machine---automatically adapting to the limited memory as required (a hallmark of scalable database engines). The other solutions failed even to perform mini-batch training on fewer than eight machines for this data set.

The ability to scale in terms of model and data size is very important, given the growing evidence that far more often than not, ``bigger is better'' in modern ML. Getting a ML system to work as embedding sizes are increased (that is, as we use ever-higher-dimensional hidden layer activations) is difficult, as this has a significant effect on memory usage. This is strong motivation for having a distributed ML system that scales with little or no human effort.

We also point out that getting these other systems to scale---even to the extent shown in the experimental results reported here---can be an arduous task.  \texttt{AliGraph} requires the user to load whole graph into memory and manually partition it for distributed training 
(\texttt{AliGraph} plans to support the METIS ~\cite{metis} graph partitioning algorithm in the near future).
\texttt{DistDGL} can partition relatively small graphs in both automatic and distributed fashion using its API \texttt{dgl.distributed.partition\_graph}. However, for large graph partitioning, a user needs to run an external tool - ParMETIS ~\cite{ParMETIS}, which involves a lot of graph format conversions. 
ParMETIS loads the graph into the memory of a single node and then sends the edges to other nodes. Arguably, the relational solution is turnkey: simply load the graph into relational tables, auto-diff the SQL, and begin training.

\section{Related Work}
Auto-differentiation has been integrated into many programming systems, including machine learning systems \cite{chen2015mxnet,TensorFlow, pytorch, jaxpaper, tangent, tokui2019chainer}, scientific computing systems \cite{bischof2003automatic, hascoet2013tapenade, slucsanschi2016adijac,DBLP:journals/corr/RevelsLP16,innes2020sense} and physical simulation systems \cite{de2018end,hu2019difftaichi,jakob2019enoki,degrave2019differentiable,heiden2021neuralsim}.
Some of the closest work to our own involves auto-diff for functional programming languages. \cite{10.1145/3341701} shows how to differentiate a higher-order functional array-processing language. \cite{10.1145/3371106} proposes a first-order language with reverse-mode differentiation. \cite{baydin2015diffsharp} adds auto-differentiation support to .NET ecosystem. \cite{pearlmutter2008reverse} incorporates auto-diff into lambda calculus. \cite{10.1145/3468791.3468840} considers auto-differentiation of the numerical kernel functions used in
RA/SQL.

Some previous work has unified RA with machine learning computations ~\cite{geerts2021matrix,zhang2021distributed,10.1145/3514221.3526150,zhou2022serving,fegaras2022scalable,guan2023comparison,rusu2023multidimensional}. ~\cite{10.14778/3467861.3467869,10.14778/3551793.3551833,park2022end,10.14778/3554821.3554853} build foundation for fusing relational operations into tensor runtime. ~\cite{kovach2023indexed} defines an intermediate representation of contraction expression for both tensor and relational computations.

One of the contributions of this work was the definition of a functional RA that can be used to form database computations on which the gradient operation can be applied. Relations in our functional RA are related to $K$-relations \cite{green2007provenance}. $K$-relations are used to build up potentially complicated computations over some set $K$, in the same way that we use RA to build computations over tensors. However, the RA defined over $K$-relations is not functional in the sense that it does not actually build functions over relations, it directly operates on them.  Hence it does not directly address the need for a functional RA.

\section{Conclusions}

We have considered the problem of automatic differentiation in relational algebra. We have demonstrated experimentally that a relational engine running an auto-diff computation can execute various ``big data'' ML tasks as fast as special-purpose distributed ML systems.  We have shown that the relational approach has the benefit that it naturally scales to very large problems, even when limited memory is available. 

\textbf{Acknowledgements.} We would like thank the anonymous reviewers for their comments on the submitted version of the paper. Work presented in this paper has
been supported by an NIH CTSA, award No. UL1TR003167 and by
the NSF under grant Nos. 1918651, 1910803, 2008240, and 2131294.
\bibliography{main}
\bibliographystyle{icml2023}
\newpage
\appendix
\onecolumn

\newpage

\section{Extensions To Arrays}
\label{sec:extensions}

In this paper, we have assumed that the \texttt{value} in each relation is a single real number.  Effectively, this assumes that the data is stored sparsely.  For example, if we wish to store a matrix \texttt{A} relationally, the sparse relational representation is:

\begin{SQL}
A(row INT, col INT, value DOUBLE)
\end{SQL}

Any missing value is assumed to be zero.
However, as mentioned previously, this can have performance degradation if a relation is used to store a vector, matrix, or higher-dimensional tensor that is not sparse.  There is a small, fixed-size cost associated with pushing each tuple through the system so a large dense computation may be problematic.  In this case, it may make more sense to store the data densely as ``chunks'':

\begin{SQL}
A(row INT, col INT, value MATRIX[][])
\end{SQL}

A dense matrix stored in this fashion, along with well-implemented, high-performance CPU or GPU kernels to operate over them, can result in excellent performance. 

Fortunately, the ideas in this paper are easily extended to such ``tensor-relational'' computations, simply by extending the kernel functions so that they operate over tensors rather than scalars.
This only requires being able to differentiate the kernel functions, which can be done by a conventional auto-diff framework such as JAX \cite{jax2018github}. By storing tensors in relations, RA auto-diff provides an automatic and efficient method to automatically generate distributed backpropagation algorithms. We provide a simple python tool can be used for RA auto-differentiation: \url{https://github.com/anonymous-repo-33/relation-algebra-autodiff}

\section{Experiment: Non-Negative Matrix Factorization}

\noindent 
\textbf{Task evaluated.}
We first benchmark a large-scale non-negative matrix factorization (NNMF).
We are given the relation $\texttt{Node}$ with schema \texttt{(ID INT, vec VECTOR[LEN])}, storing node identifiers and embeddings of the nodes in the graph, and $\texttt{Edge}$, which is a relation storing all the edges in a graph.
The total number of nodes is $N$. 
The dimensionality of the node embedding is $D$.  We run experiments with the following four cases:
(1) $N = 40$k, $D = 40$k;
(2) $N = 50$k, $D = 40$k;
(3) $N = 60$k, $D = 10$k;
(4) $N = 10$k, $D = 60$k.

\noindent 
\textbf{Experiments.}
We benchmark the RA implementation (\texttt{RA-NNMF}) against \texttt{Dask}~\cite{Dask}, a popular parallel computation framework, and 
a careful ``by-hand'' implementation on top of MPI.
All three implementations are using stochastic gradient descent (SGD) with learning rate $\eta = 0.1$.
Node embeddings are randomly initialized.

\begin{figure}[ht]
    \centering
    \vspace*{-1.7cm}
    \subfigure{
    \label{fig:nnmf-1}
    \includegraphics[width=0.23\textwidth]{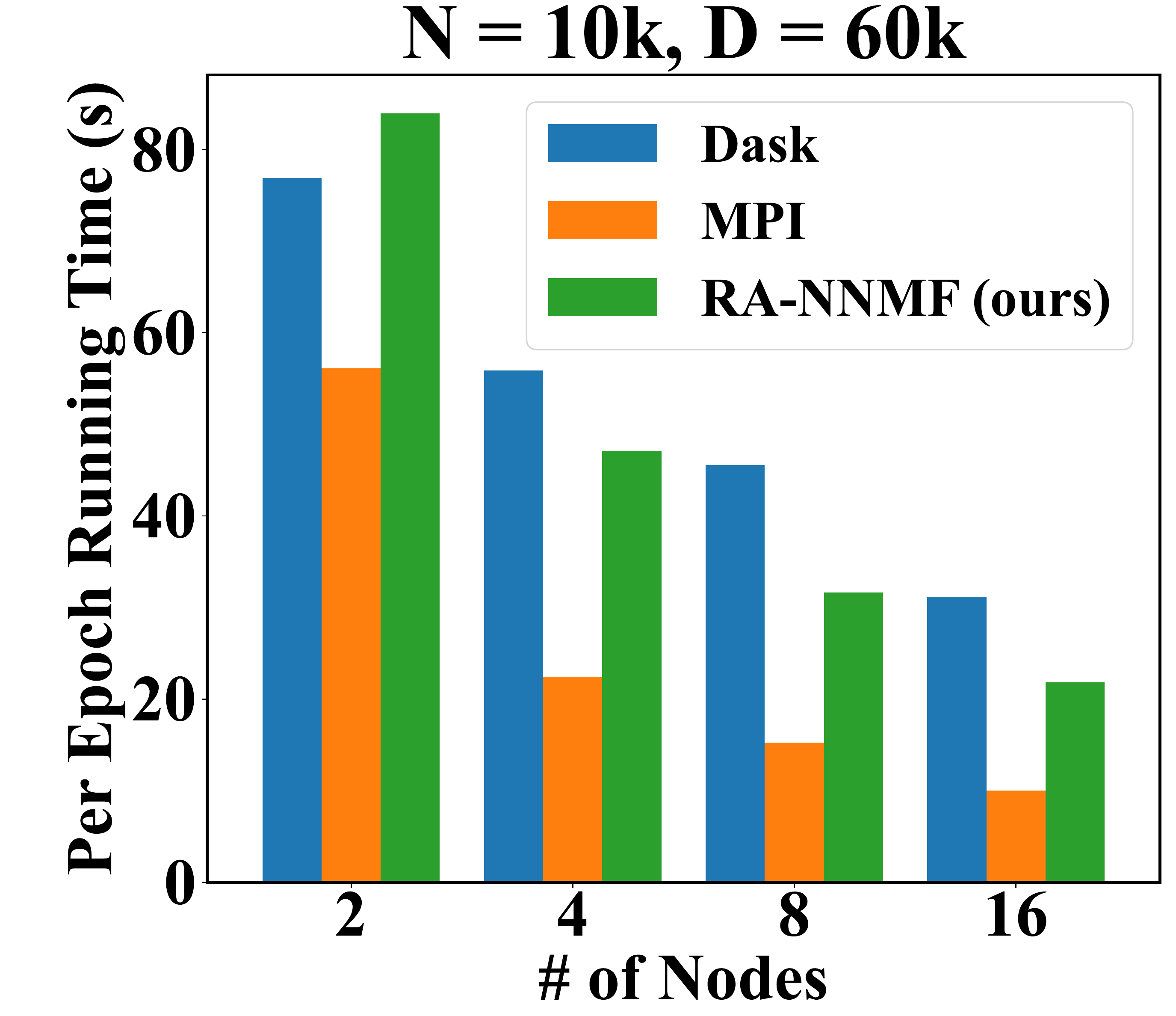}}
    \subfigure{
    \label{fig:nnmf-2}
    \includegraphics[width=0.23\textwidth]{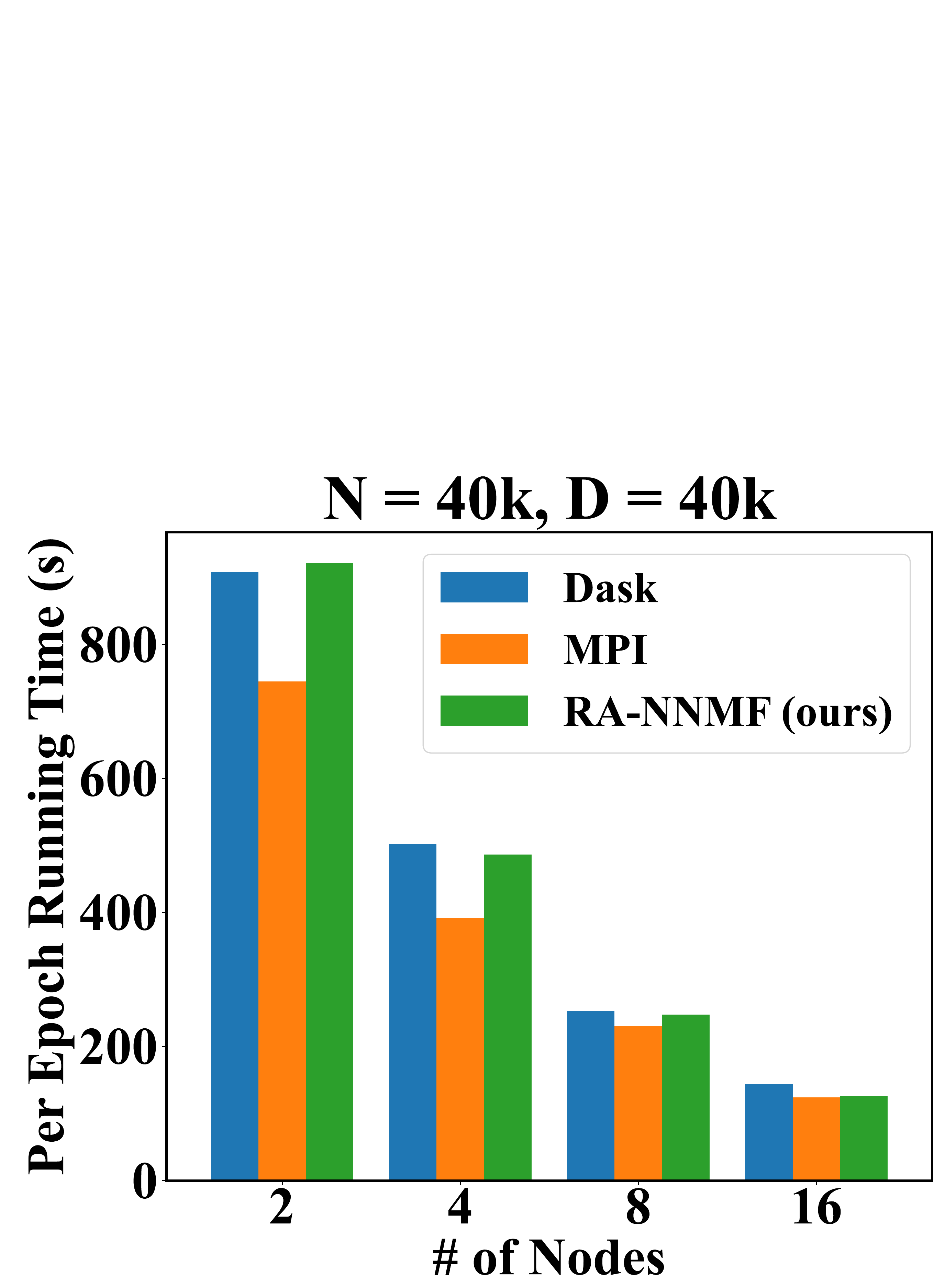}}
    \subfigure{
    \label{fig:nnmf-3}
    \includegraphics[width=0.23\textwidth]{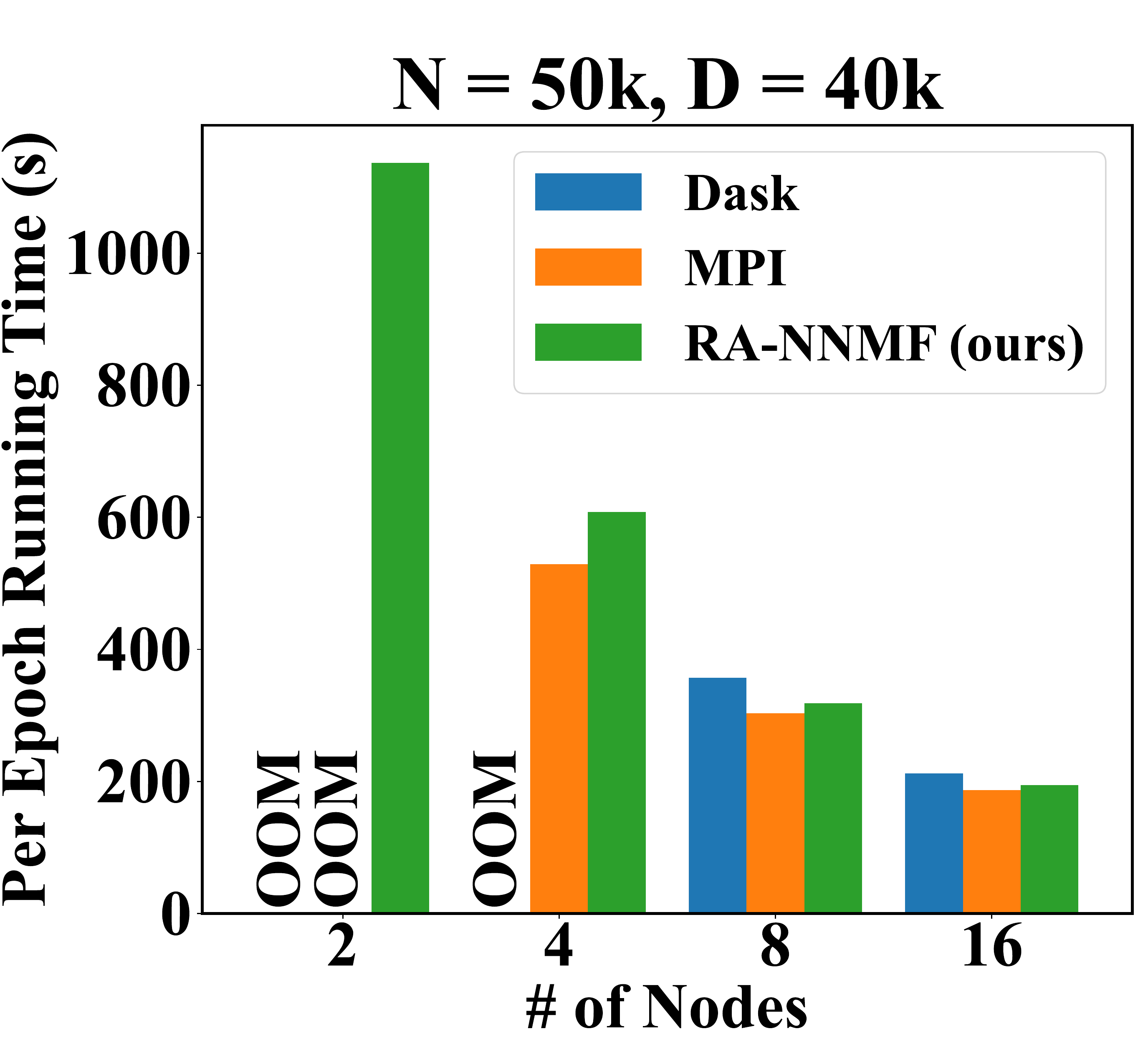}}
    \subfigure{
    \label{fig:nnmf-4}
    \includegraphics[width=0.23\textwidth]{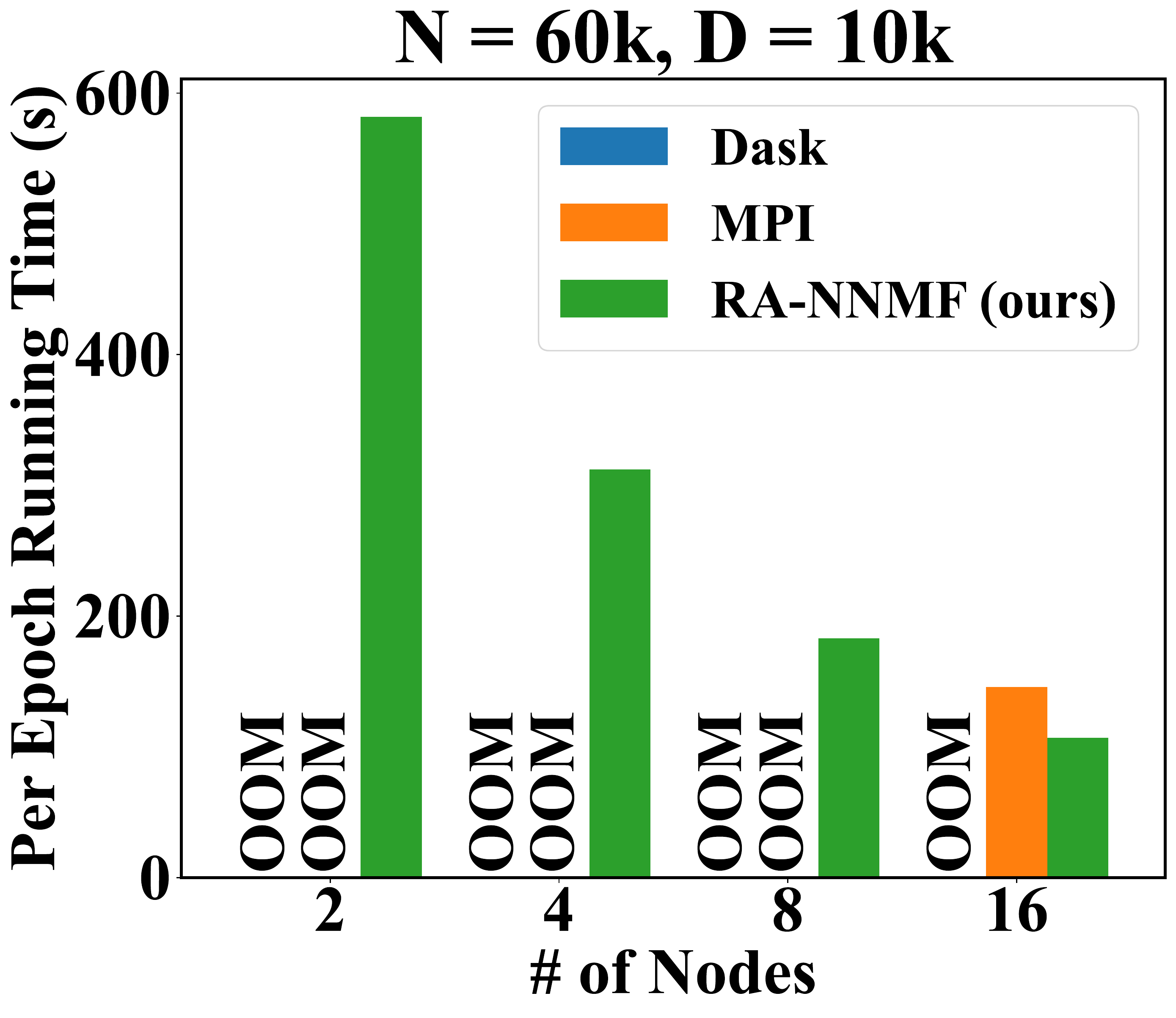}}
    \caption{NNMF per-epoch running times.}
    \label{fig:nnmf}
\end{figure}

\noindent 
\textbf{Results.}
We record per-epoch running time of three implementations in different cluster sizes: 2, 4, 8, and 16.
The results are shown in Figure ~\ref{fig:nnmf}.
\texttt{Dask} heavily relies on the large memory capacity of the clusters and runs out of memory (OOM) during backward propagation for the case $N=60$k, $D=10$k.

\section{Experiment: Knowledge Graph Embedding}
\noindent 
\textbf{Task Evaluated.}
Finally, we implement two common knowledge graph embedding (KGE) algorithms: TransE-L2 \cite{TransE} and TransR \cite{TransR}.
\noindent 
\newpage
\begin{figure}[ht]
    \centering
    \subfigure{
    \label{fig:kge-1}
    \includegraphics[width=0.3\textwidth]{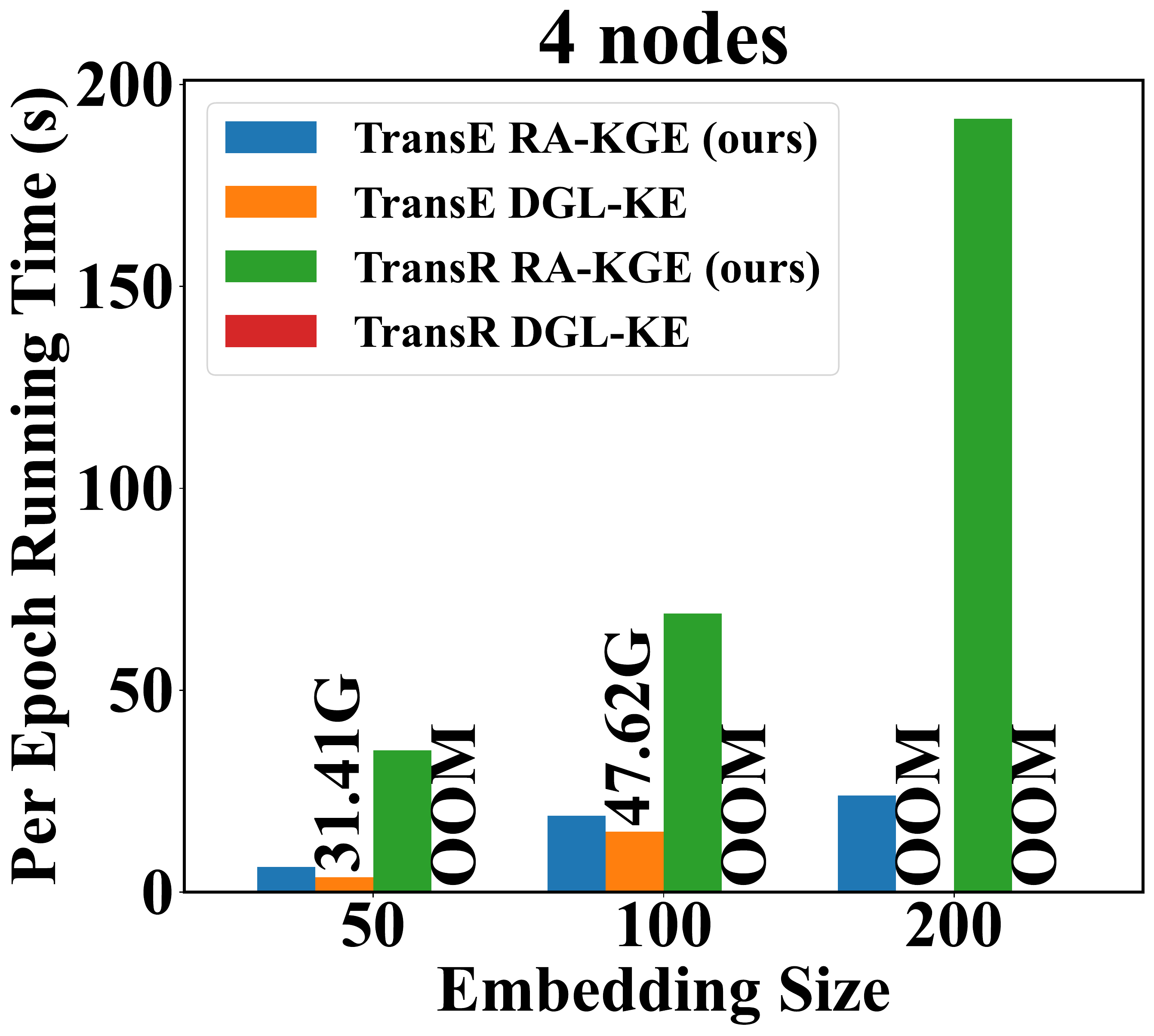}}
    \subfigure{
    \label{fig:kge-2}
    \includegraphics[width=0.3\textwidth]{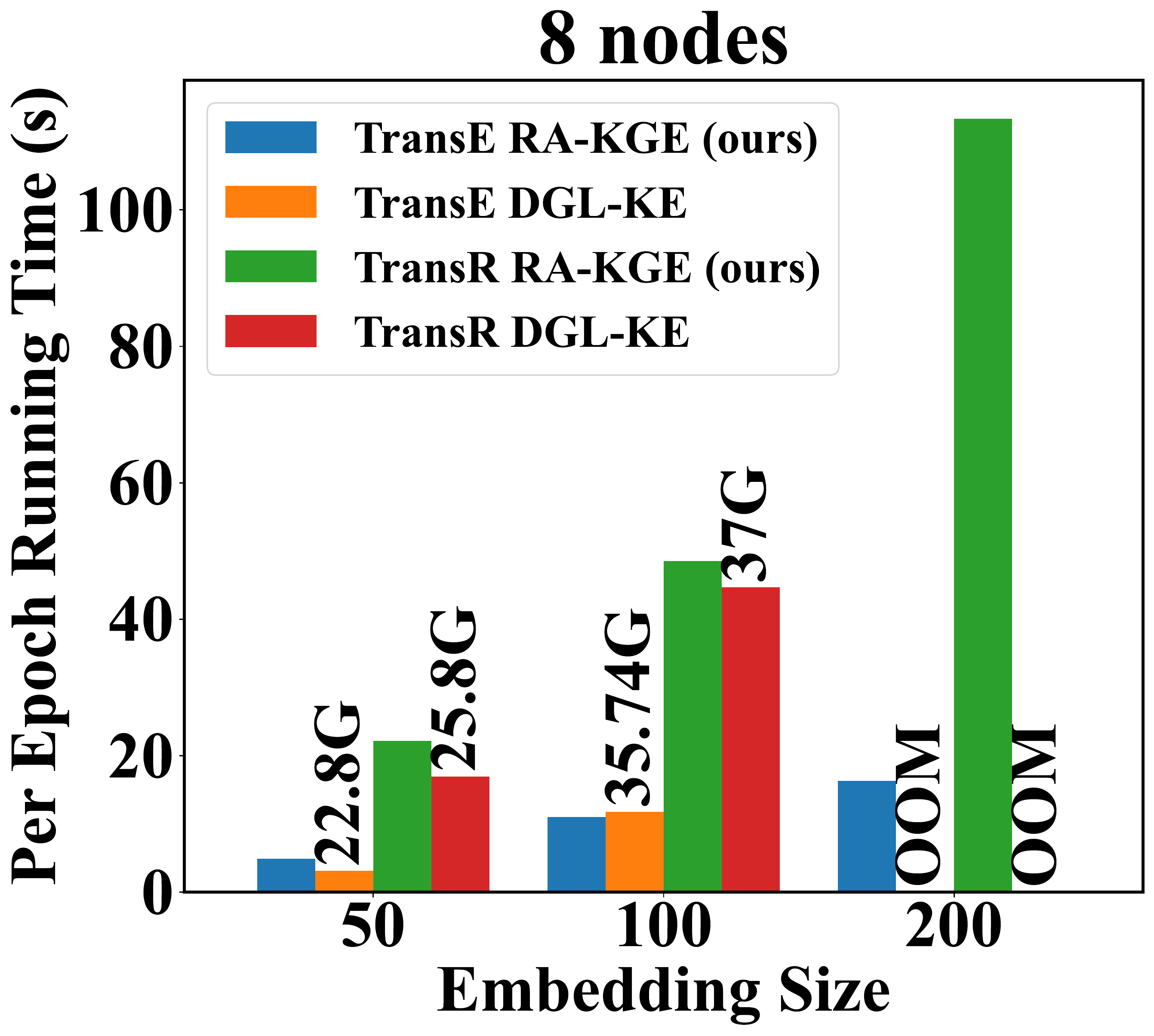}}
    \subfigure{
    \label{fig:kge-3}
    \includegraphics[width=0.3\textwidth]{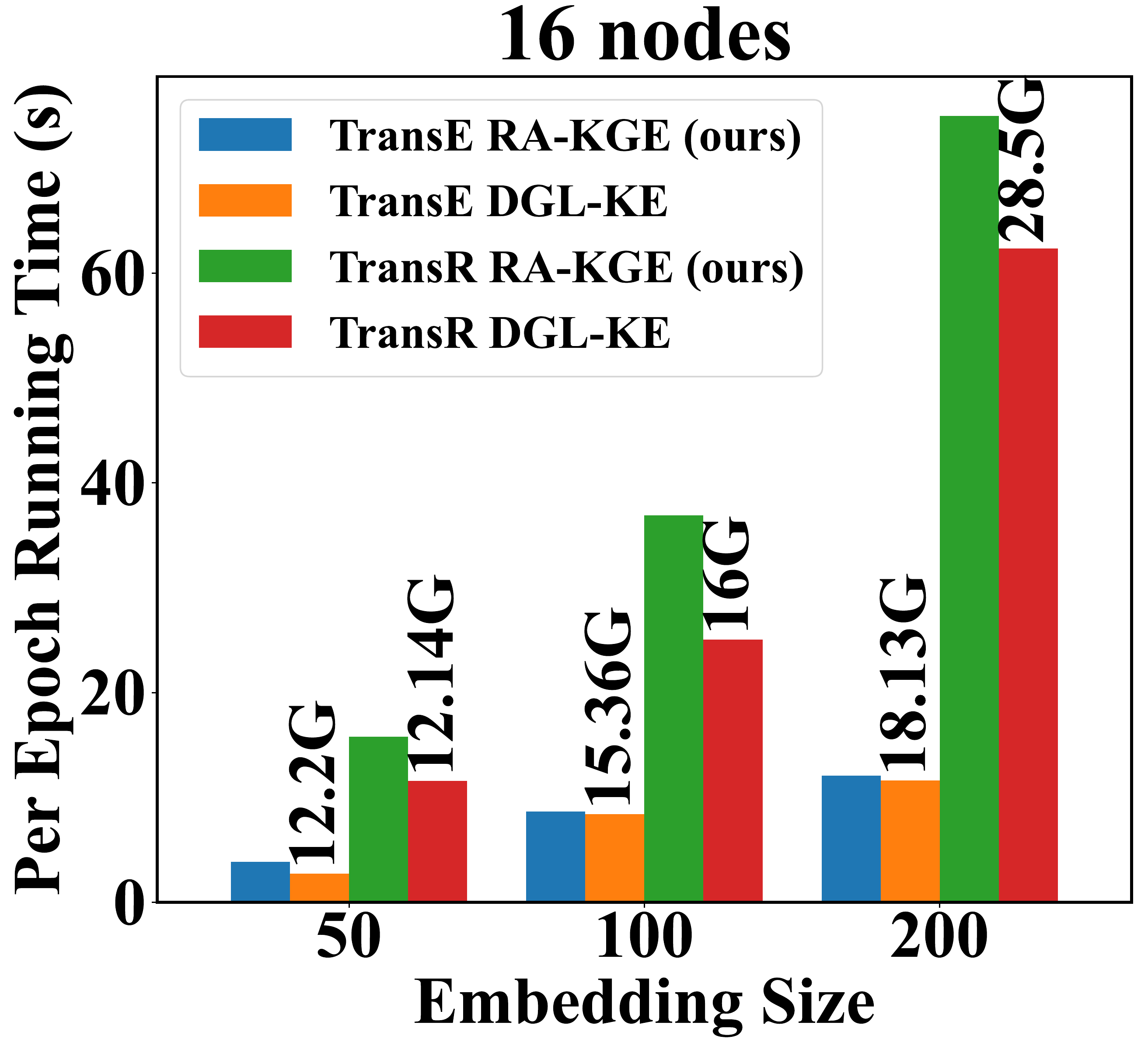}}
    \caption{100-iteration time for knowledge-graph-embedding training on Freebase; batch size is 1K.}
    \label{fig:kge}
\end{figure}
\textbf{Experiments.}
We train our KGE model on the \texttt{Freebase} data set. \texttt{Freebase} ~\cite{freebase} contains 1.9 billion triples in RDF format; it is a knowledge graph with 86M nodes, 339M edges, and 14,824 relations.
We refer to our PlinyCompute-based RA implementation (auto-generated via our relational auto-diff) as \texttt{RA-KGE}.
We compare against the distributed knowledge graph embedding training framework \texttt{DGL-KE} ~\cite{DGL-KE}.
We split the dataset into a training set (90\%), a validation set (5\%), and a testing set (5\%).
For each positive sample, 200 corrupted negative samples are used.
We pick the entity embedding size $D = 50, 100, 200$;
For TransE, we choose the same embedding size for both relations and entities.
For TransR, we choose the double entity embedding size for relations.
The optimizer is SGD with learning rate $\eta = 0.5$.
We consider three different cluster sizes: 4, 8, and 16 nodes.
For \texttt{DGL-KE}, the dataset is manually partitioned into 4, 8, and 16 parts using METIS.

\vspace{1 pt}
\noindent 
\textbf{Results.} We observe and compare the time to perform 100 forward and back-prop iterations for each of the various experimental settings.
The results are shown in Figure ~\ref{fig:kge}.  For \texttt{DGL-KE} the number after the per-iteration running time is the maximum per-node memory usage.  OOM is reported if the system failed due to lack of memory.
\newpage
\section{Example for RJPs}
\begin{figure}[h]
  \centering
  \includegraphics[width=\textwidth]{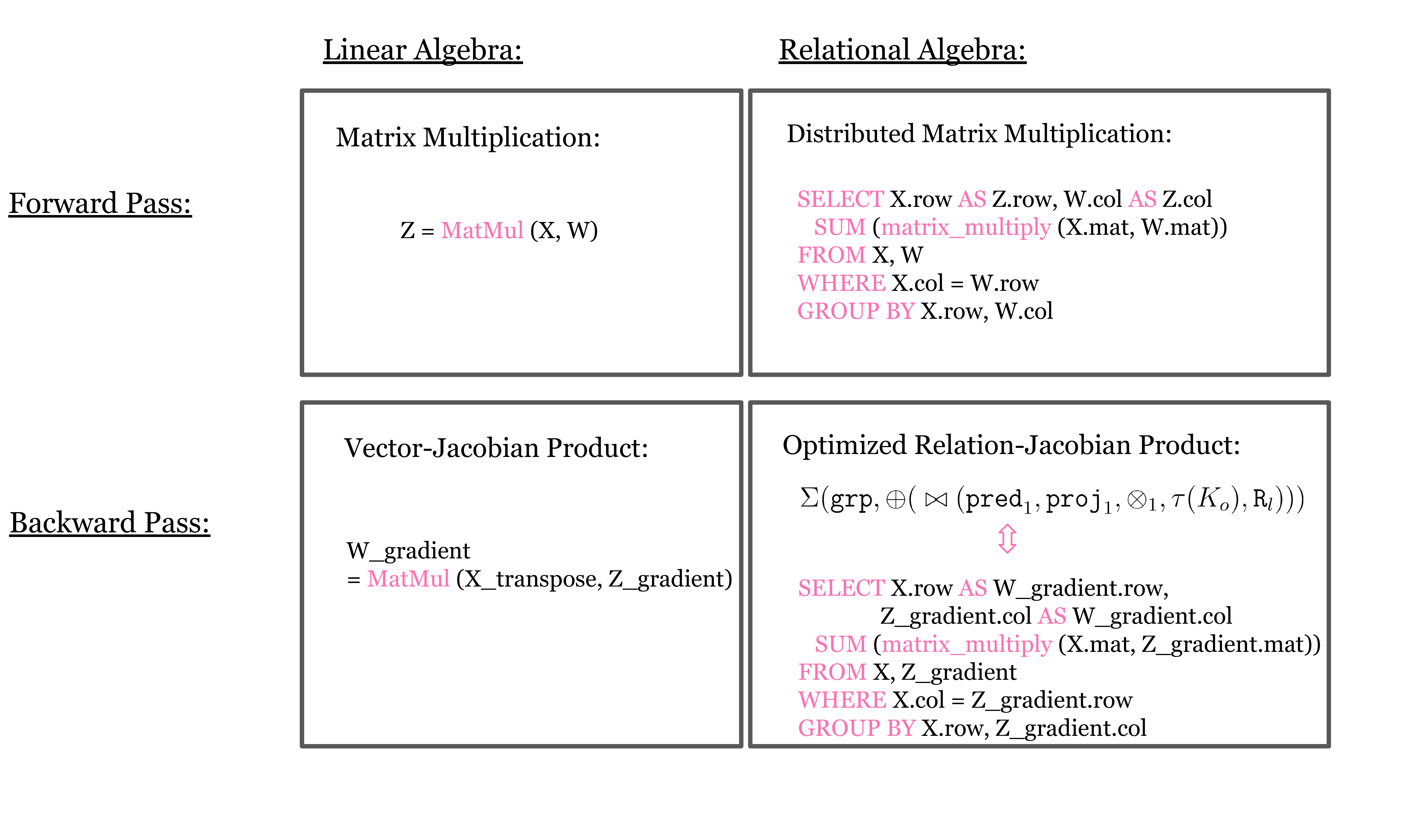}
  \vspace{-10 pt}
  \caption{Correspondence between the same computation in linear algebra (left) and relational algebra/SQL (right) in forward pass for computing Z and backward pass for computing gradients of W. The input matrices X and W are stored into relations by decomposing matrices into chunks or blocks and operated over using high-performance kernels (such as matrix\_multiply). The RA-based computation generated via auto-diff executed on a high-performance database engine provides an easy way to run a distributed backpropagation algorithm \cite{yoon1990distributed}.}
  \label{fig:compare-LA-RA-autodiff}
\end{figure}
\begin{figure}[!]
  \centering
  \vspace*{-5cm}
  \includegraphics[width=0.6\textwidth]{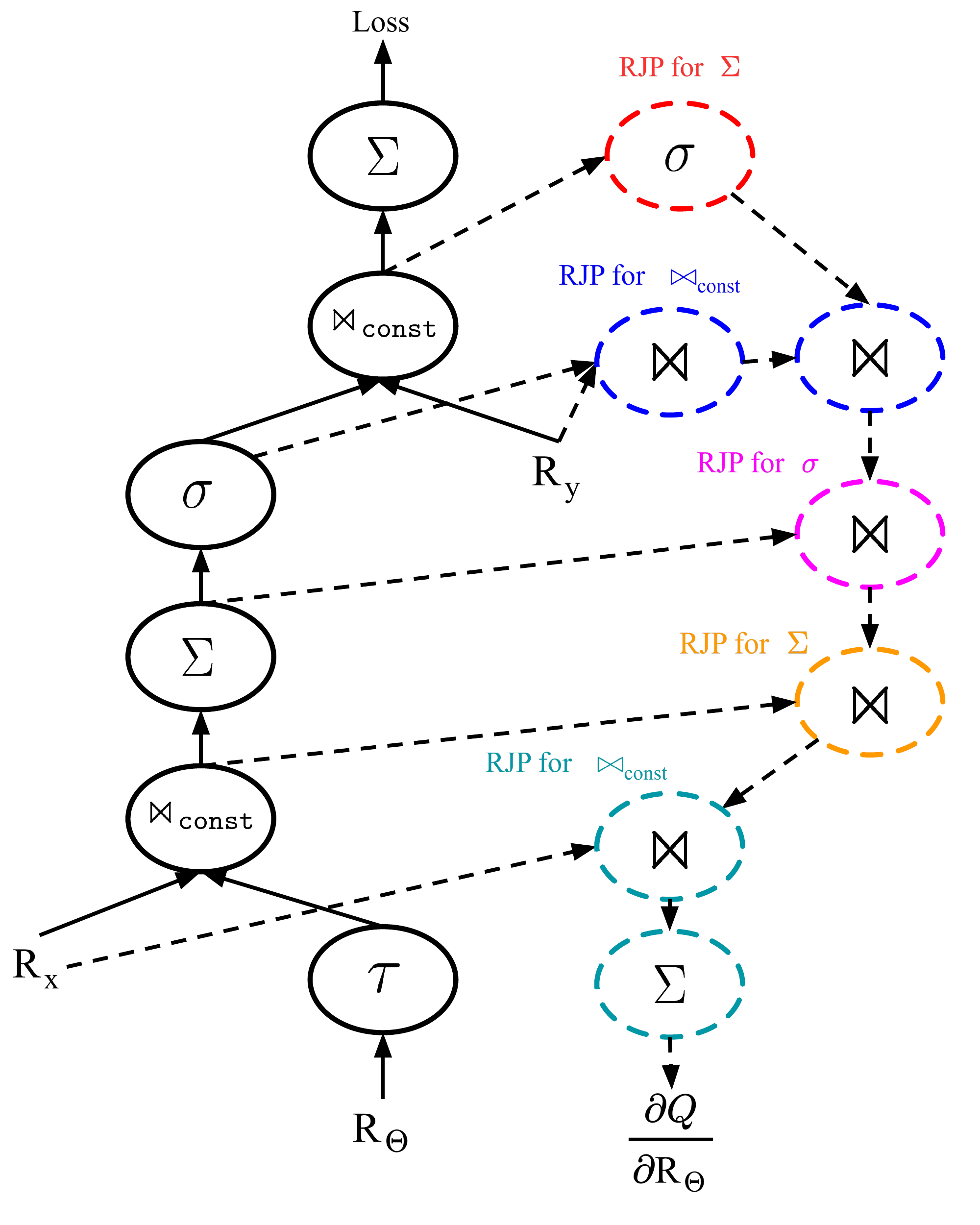}
  \caption{The left part is the logistic regression. The right part is the generated query by RJPs for differentiating parameters in logistic regression. The top $\Join_\texttt{const}$ is a $\Join_{1-1}$ while the bottom $\Join_\texttt{const}$ is a $\Join_{1-n}$. We apply all RJP optimizations for $\Sigma$ and $\Join$ mentioned in Section~\ref{sec:RJPs}.}
  \label{fig:autodiff}
\end{figure}

\end{document}